%% file: iclr2027_conference.tex
\documentclass{article} %
\usepackage{iclr2027_conference,times}

\usepackage{hyperref}
\usepackage{url}

\usepackage[utf8]{inputenc} %
\usepackage[T1]{fontenc}    %

\iclrfinalcopy

\usepackage{url}            %
\usepackage{booktabs}       %
\usepackage{amsfonts}       %
\usepackage{nicefrac}       %
\usepackage{microtype}      %
\usepackage{xcolor}         %

\usepackage{amsmath}
\usepackage{amssymb}
\usepackage{mathtools}
\usepackage{amsthm}
\usepackage{multirow}
\usepackage[table]{xcolor}
\usepackage{microtype}
\usepackage{graphicx}
\usepackage{subcaption}
\usepackage{booktabs}
\usepackage{pifont}
\usepackage{float}
\usepackage{caption}
\usepackage{wrapfig}
\usepackage{algorithm}
\usepackage{algorithmic}

\input{macro}
\newcommand{\cmark}{\ding{51}} %
\newcommand{\xmark}{\ding{55}} %
\definecolor{hl}{gray}{0.85}

\theoremstyle{plain}
\newtheorem{theorem}{Theorem}[section]

\theoremstyle{definition}

\theoremstyle{remark}

\title{Sample-wise Targeted Adversarial Attacks on Test-time Adaptation}

\author{%
  Phuc Duc Nguyen \\
  College of Computing and Data Science\\
  Nanyang Technological University\\
  Singapore\\
  \texttt{ducphuc001@e.ntu.edu.sg} \\
  \And
  Quang Duc Nguyen \\
  College of Computing and Data Science\\
  Nanyang Technological University\\
  Singapore\\
  \texttt{quangduc002@e.ntu.edu.sg} \\
}

\begin{document}

\maketitle
\lhead{}

\input{0_abstract}

\input{1_introduction}

\input{2_related_works}

\input{3_motivation}

\input{4_method}
\input{5_experiment}

\input{7_conclusion}

\subsection*{AI use statement}

We used generative AI tools primarily to assist with manuscript writing, including grammar correction, language polishing, improving clarity, and refining the presentation of existing technical content. We also used generative AI to assist with generating and debugging well-defined sections of experimental code for which the intended functionality, inputs, and outputs were specified by the authors. All AI-assisted text and code were reviewed, edited, and verified by the authors before inclusion in the final work.

\subsection*{Ethics statement}
This work investigates security vulnerabilities in Test-Time Adaptation (TTA) systems and therefore has potential dual-use implications. The proposed attack is developed solely for research purposes and is evaluated in controlled experimental settings using standard benchmark datasets and models; we do not target real-world deployed systems or use sensitive user data. While the methodology could potentially be misused to compromise adaptive systems, exposing such vulnerabilities is important for understanding the security risks of deployment-time learning and for motivating more robust defenses. We therefore present the attack to support responsible security research and the development of safer TTA mechanisms. To the best of our knowledge, this work does not involve human subjects, private data, or other ethical concerns beyond the security and dual-use considerations described above.

\subsection*{Reproducibility statement}

We provide detailed experimental settings and implementation details in the main paper and appendices, including the attack configuration, hyperparameters, TTA settings, baseline adaptations, trigger specifications, and evaluation protocols. The official implementation code will be made publicly available upon acceptance.

\bibliography{reference}
\bibliographystyle{iclr2027_conference}

\input{8_appendix}

\end{document}

%% file: 0_abstract.tex
\begin{abstract}
Test-time adaptation (TTA) mitigates distribution shifts by adapting models to unlabeled test inputs, but also exposes them to adversarial manipulation. Existing class-wise targeted attacks remain suboptimal for stealthy exploitation in this setting: since TTA operates on batches, forcing a subset of samples toward a target label unintentionally pulls similar benign samples along, resulting in a conspicuously high frequency of the target label that is easy to detect. 
To capture a more realistic threat, we introduce a sample-wise targeted attack. Unlike prior approaches, the attacker aims to misclassify only inputs carrying an attacker-chosen trigger, while preserving a benign-like prediction distribution to evade detection. To achieve this, we propose a meta-learning-based attack with a novel priority-aware gradient alignment strategy that explicitly prioritizes attack success. The strategy formulates the gradient update as an ellipsoidal trust-region problem, mitigating gradient conflict between the attack and stealth objectives, while providing theoretical guarantees for effective optimization of the attack objective in the presence of gradient misalignment. Extensive experiments on CIFAR-10-C, CIFAR-100-C, and ImageNet-C across TTA protocols demonstrate that our method achieves high targeted success rates while maintaining prediction behavior close to benign adaptation under multiple label-free stealth metrics, making it difficult to detect in unlabeled TTA deployment scenarios. Furthermore, we demonstrate that our attack shows strong robustness against existing defenses.
\end{abstract}

%% file: 1_introduction.tex
\section{Introduction} \label{sec:intro}

Deep models deployed in the wild inevitably face distribution shift between training and deployment data due to changes in sensors, environments, or usage patterns, which can severely degrade performance even for models that are robust on in-distribution benchmarks \cite{hendrycks2019benchmarking}. Training-time remedies such as domain adaptation \cite{li2024comprehensive} and robust training \cite{qian2022survey} mitigate some shifts but cannot anticipate all future conditions and often require source data and labels that are unavailable at deployment. Test-time adaptation (TTA) addresses this gap by updating model parameters online using unlabeled test inputs \cite{liang2025comprehensive}: by minimizing entropy \cite{wang2020tent, lee2022surgical}, performing self-training \cite{su2022revisiting,sinha2023test}, or enforcing consistency on the incoming test stream \cite{wang2022continual, brahma2023probabilistic}. Without the training set, TTA can effectively recover performance under distribution shift.

\begin{figure*}[t]
\begin{minipage}[t]{0.52\linewidth}
    \includegraphics[width=\linewidth]{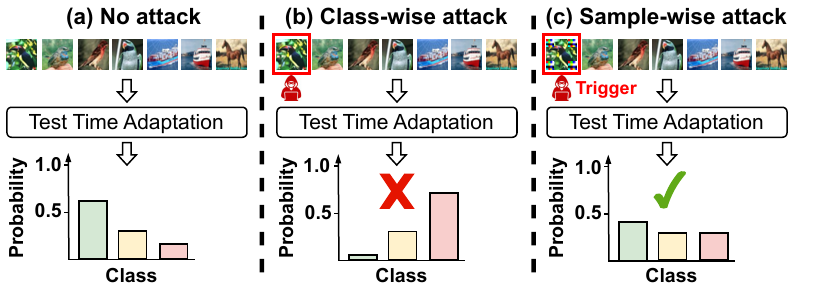}
    \vspace{-0.5cm}
    \captionof{figure}{Illustration of distributional stealth. Unlike class-wise attacks \textbf{(b)} that create detectable anomalies by indiscriminately affecting \emph{all} samples in a class, the proposed sample-wise attack \textbf{(c)} uses a specific trigger to misclassify \emph{only} the intended victim, preserving a natural, stealthy prediction distribution to evade detection.}
    \label{fig:example}
\end{minipage}
\hfill
\begin{minipage}[t]{0.47\linewidth}
    \Huge
    \centering
    \vspace{-2.5em}
    \captionof{table}{Taxonomy of existing adversarial threat models under TTA. Unlike prior works that only enable class-wise targeted attacks, our threat model allows an adversary to selectively induce misclassification on specific inputs, possibly from different classes (sample-wise).}
    \label{tab:taxo}
    \resizebox{\linewidth}{!}{
    \begin{tabular}{ccccc}
    \toprule
    {\textbf{Threat}} &
    {\textbf{Model}} &
    {\textbf{Targeted}} &
    \multirow{2}{*}{\textbf{Selectivity}} &
    \textbf{Access to} \\
    \textbf{model} & \textbf{Knowledge} & \textbf{Attack} & & \textbf{user data} \\
    \midrule
    DIA \cite{wu2023uncovering} & White-box & \cmark & Global/ class-wise   & \cmark  \\
    TePA \cite{cong2024test} & Grey-box & \xmark & Global & \xmark  \\
    FCA \cite{rifatadversarial} & Grey-box & \cmark & Global/ class-wise   & \xmark \\
    RTTDP \cite{su2024adversarial} & Grey-box & \cmark & Global/ class-wise   & \xmark \\
    \midrule
    Ours (this work) & Grey-box & \cmark & \textbf{Sample-wise} & \xmark  \\
    \bottomrule
    \end{tabular}
    }
\end{minipage}
\vspace{-2em}
\end{figure*}

However, because TTA implicitly trusts the unlabeled test stream to drive online updates, it introduces a new attack surface \cite{wu2023uncovering}. By manipulating a subset of the test samples used for a single update, an adversary can influence predictions on other inputs processed in the same batch by a shared TTA model \cite{su2024adversarial}. Existing literature largely focuses on two threat models. The first is \emph{untargeted} attack, in which the adversary drives the model into broadly poor performance, offering limited concrete benefit to a rational attacker. The second is \emph{class-wise targeted} attack \cite{wu2023uncovering,cong2024test,su2024adversarial,rifatadversarial}. The latter, while more strategic, remains suboptimal: by indiscriminately biasing predictions for all samples resembling the victim class, it causes widespread, easily detectable anomalies (e.g., constant prediction of the target class). Since TTA has no access to ground truth labels during deployment, direct accuracy monitoring is infeasible. Thus, we consider \emph{distributional stealth}, i.e., maintaining label-free prediction statistics close to a benign baseline, as a practical proxy for stealth, motivated by label-shift detection via prediction distributions~\cite{lipton2018detecting} and evidence that marginal distribution shifts are effective drift signals~\cite{rabanser2019failing}. 
As illustrated in Figure~\ref{fig:example}, \emph{class-wise targeted} attacks are unstealthy as they induce an unnatural concentration of predictions on the target, which could trigger anomaly alarms, especially when the victim class appears frequently. In contrast, realistic adversaries often prioritize selective manipulation to avoid detection. Therefore, it is important to investigate a finer-grained threat model that explicitly captures the selective nature of attacks under TTA, to guide the development of more robust TTA defenses.

We study \emph{sample-wise targeted} attacks on TTA, in which only inputs carrying an attacker-chosen trigger (possibly from different classes) should be redirected to a target label, while non-triggered inputs from any user preserve a benign-like prediction distribution to evade detection. Instantiating this threat model presents two key challenges. First, enforcing sample-wise selectivity under realistic TTA constraints is difficult, as the attacker has no access to benign or victim user data and must rely solely on attacker-controlled samples~\cite{su2024adversarial}. Second, existing class-wise attacks~\cite{wu2023uncovering, su2024adversarial, rifatadversarial} do not consider trigger information during optimization; consequently, naively applying them to the sample-wise setting yields poor performance. 

To overcome these limitations, we propose a meta-learning-based attack that simulates the adaptation process to best utilize the attacker's limited data. This framework optimizes the adaptation process with two objectives: \textbf{\emph{(i)}} a primary attack objective that maximizes the success rate on trigger-carrying inputs, and \textbf{\emph{(ii)}} a secondary distributional stealth objective that prevents the conspicuous output anomalies. We introduce a novel loss term for \textbf{\emph{(ii)}}, tailored to the TTA setting. However, we identify a critical optimization challenge: the gradients of these two objectives are strongly antagonistic. 
Standard multi-task formulations \cite{sener2018multi,yu2020gradient,liu2021conflict} fail here because they treat tasks symmetrically, seeking Pareto compromises that balance competing objectives. In an adversarial context, however, such symmetry is detrimental: it often sacrifices the attack effectiveness for marginal gains in stealth. Since a failed attack has zero utility to the adversary regardless of its stealthiness, the objectives are inherently \emph{asymmetric} in our setting, where attack success is primary, while stealth functions as a constraint. To enforce this, we propose a novel \emph{priority-aware gradient alignment} formulated via an ellipsoidal trust-region. Our update direction is constructed to stay close to the attack gradient while penalizing movement toward misaligned stealth-gradient directions, yielding updates that are attack-driven yet distributionally consistent. We provide theoretical guarantees showing that this aligned direction ensures descent on the attack objective for the unprojected update even under severe gradient misalignment, whereas symmetric multi-objective baselines lack such guarantees and may fail to make progress. Finally, we empirically demonstrate that our method achieves a high sample-wise targeted attack success rate of around 90\% while maintaining label-free prediction statistics close to the no-attack baseline. Contributions of this paper are summarized as follows:
\begin{itemize}
    \item We formalize a sample-wise targeted threat model for TTA that enforces strict selectivity: attacking only trigger-carrying inputs while maintaining a benign-like prediction behavior for non-triggered queries.
    \item We propose a meta-learning-based attack with a novel priority-aware gradient alignment that mitigates the gradient misalignment between attack objectives and distributional stealth, with a descent guarantee for the unprojected update and a sufficient descent condition under projection.
    \item Extensive experiments demonstrate that our method achieves high targeted attack success rates while maintaining benign-like behavior under multiple label-free stealth metrics.
\end{itemize}

%% file: 2_related_works.tex
\vspace{-0.2cm}
\section{Related Works}
\vspace{-0.2cm}
We discuss the related work of this study below, with additional discussion provided in Appendix~\ref{appendix:more_rw}.
\vspace{-0.8em}
\paragraph{Test-time adaptation and attacks.} TTA handles distribution shifts by updating model parameters on unlabeled test streams, typically via entropy minimization \cite{wang2020tent,zhang2024come} or self-training \cite{niu2023towards,rusak2021if}. While effective, this mechanism implicitly trusts the incoming data, exposing a critical attack surface. Prior adversaries largely focus on \emph{untargeted} \cite{cong2024test} or \emph{class-wise targeted} attacks \cite{wu2023uncovering,su2024adversarial,rifatadversarial}. Both render the attack unstealthy. In contrast, we propose a sample-wise targeted threat model: misclassifying only trigger-carrying inputs while maintaining a benign-like prediction behavior. The key differences are summarized in Table~\ref{tab:taxo}.
\vspace{-0.8em}
\paragraph{Connection to backdoor attacks.} Our use of input triggers parallels traditional backdoor attacks~\cite{gu2017badnets,li2022backdoor}. However, unlike standard backdoor attacks that require poisoning training data or model parameters during training, our attack operates strictly at test time. We exploit the TTA process to induce backdoor-like behavior in clean, pre-trained models on the fly, without any prior access to either the training phase or the training data.
\vspace{-0.8em}
\paragraph{Gradient conflict resolution.} Balancing the attack and stealth objectives resembles Multi-Task Learning (MTL), where conflicting or misaligned gradients often destabilize optimization \cite{zhang2021survey}. However, unlike standard MTL solvers (e.g., \cite{yu2020gradient,liu2021conflict}) that seek symmetric trade-offs or Pareto optimality, our setting is priority-asymmetric: maximizing attack success is the primary goal, while stealth is a secondary objective. Consequently, we propose a priority-aware alignment strategy that penalizes movement toward increasingly misaligned stealth directions, rather than balancing both objectives equally.

%% file: 3_motivation.tex
\vspace{-0.2cm}
\section{Threat Model} \label{sec:threat_model}
\vspace{-0.2cm}

\begin{wrapfigure}{r}{0.5\linewidth}
\centering
\vspace{-1.3em}
\includegraphics[width=0.99\linewidth]{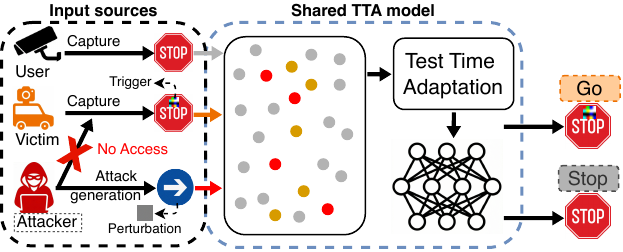}
\caption{Attack setting under shared TTA. The attack is selective, where it forces trigger-carrying victim inputs (red stream) to a target label while maintaining benign-like prediction behavior on non-triggered inputs (gray stream). Trigger injection occurs upstream or in the physical environment, and is naturally captured by victim inputs. The attacker has no access to victim data.}
\vspace{-2.5em}
\label{fig:overall_l}
\end{wrapfigure}

We consider a shared TTA service that adapts on batches containing inputs from multiple sources, including an attacker-controlled subset, as shown in Figure~\ref{fig:overall_l}. These inputs jointly influence adaptation and the predictions returned for the batch.
The same shared-deployment scenario is also considered in prior TTA attack work \cite{wu2023uncovering,cong2024test,su2024adversarial,rifatadversarial}. The core difference in our setting lies in the adversary goal, i.e., class-wise vs.\ sample-wise targeting. We next formalize the system model, adversary goals, and adversary capabilities.
\vspace{-0.1cm}
\subsection{System Model}
\vspace{-0.1cm}
Let $\mathcal{X} \subset \mathbb{R}^D$ denote the input space and $\mathcal{Y}=\{1,\dots,C\}$ the label space. We consider a multi-class classification service that deploys a TTA model $f_{\theta_t}: \mathcal{X} \to \mathbb{R}^C$, where $\theta_t$ represents the model parameter at adaptation step $t$. $\theta_0$ denotes the original pre-trained parameters prior to any adaptation. During deployment, each incoming test batches ${X}^{(t)}$ comprises of three disjoint subsets: benign inputs ${X}^{(t)}_B := \{{x}^{(t)}_{B,i}\}_{i=1}^{N_B}$, victim inputs ${X}^{(t)}_V := \{{x}^{(t)}_{V,i}\}_{i=1}^{N_V}$ (some of which may carry the attacker’s trigger), and attacker-controlled inputs ${X}^{(t)}_A := \{{x}^{(t)}_{A,i}\}_{i=1}^{N_A}$. These subsets identify the roles of inputs in the attack, rather than requiring distinct accounts or devices. All inputs lie in the same input space $\mathcal{X}$ and label space $\mathcal{Y}$, with
${X}^{(t)} = {X}^{(t)}_B \cup {X}^{(t)}_V \cup {X}^{(t)}_A$.
For every batch ${X}^{(t)}$, the service internally updates the model parameters via a TTA mechanism $\mathcal{A}$, i.e., $\theta_{t+1} = \mathcal{A}(\theta_t, {X}^{(t)})$, and subsequently outputs predictions $f_{\theta_{t+1}}({X}^{(t)})$, without access to ground truth labels \cite{su2024adversarial, rifatadversarial}. 
\vspace{-0.1cm}
\subsection{Adversary Goals} \label{sec:adv_goal}
\vspace{-0.1cm}
The objective of the \emph{sample-wise targeted} attack on TTA is twofold. First, it seeks to induce {\textbf{targeted misclassification on a selected subset}}, ensuring that a predefined portion of victim inputs $\hat{X}^{(t)}_v \subseteq {X}^{(t)}_V$ is mapped to a specific target label ${y}_{\mathrm{tgt}} \in \mathcal{Y}$ after adaptation. Second, since direct accuracy monitoring is infeasible due to the absence of ground truth labels during deployment of TTA, the attack prioritizes {\textbf{stealth via label-free prediction consistency}}, requiring that the adapted model maintains benign-like prediction behavior on all remaining non-targeted inputs, i.e., ${X}^{(t)}_B \cup \bigl({X}^{(t)}_V \setminus \hat{X}^{(t)}_v\bigr)$, thereby rendering the attack difficult to detect.
\vspace{-0.8em}
\paragraph{Trigger-defined victim subset.}
Rather than selecting victims by their semantic class, we assume the attacker defines $\hat{X}^{(t)}_v$ via the presence of a \emph{trigger}, adopting the formalism from backdoor attacks~\cite{gu2017badnets,chen2017targeted}. Let $T:\mathcal{X}\!\rightarrow\!\mathcal{X}$ denote a trigger operator (e.g., a localized patch or a sinusoidal pattern), and define the triggered victim set $\hat{X}^{(t)}_v := \{\,T(x)\mid x\in X^{(t)}_v\,\}$, where ${X}^{(t)}_v$ is the selected untriggered set. Importantly, trigger injection occurs \emph{outside} the shared TTA pipeline: it may be physical (e.g., a patch on an object captured by sensors) or upstream/digital (e.g., a marker in attacker-provided content processed by a victim application). In either case, the attacker controls the trigger pattern without observing or accessing the resulting victim inputs (Figure~\ref{fig:overall_l}). The attack succeeds if, after the TTA update, all triggered victim inputs are forced to the same target label, i.e., $\forall {x} \in {\hat{X}}^{(t)}_v: f_{\theta_{t+1}}\left({x}\right) = {y}_{\mathrm{tgt}}$, while prediction behavior on non-triggered inputs remains close to the benign baseline.
\vspace{-0.8em}
\paragraph{Key distinction from class-wise targeted attacks.}
Prior class-wise targeted attacks couple the misbehavior to a specific source class (e.g., all ``stop" sign $\rightarrow$ ``go" sign), which inevitably distorts the global output into a degenerate probability spike. In contrast, our victim set is trigger-defined and thus independent of the underlying semantic class. This selectivity gives the attacker greater flexibility (e.g., affecting samples from multiple classes simultaneously) while allowing the adapted model to maintain benign-like prediction behavior on non-triggered inputs. In the absence of labels, this makes the attack harder to distinguish from natural adaptation.
\vspace{-0.1cm}
\subsection{Adversary Capabilities}
\vspace{-0.1cm}

The attacker can submit its own inputs to the shared TTA service but cannot observe or modify other users’ submitted inputs. The attack operates when attacker-controlled and victim inputs are processed in the same adaptation batch. We evaluate varying attacker participation in Appendix~\ref{appendix:random_participation}.
We assume a \emph{grey-box} setting, following realistic attack scenarios in \cite{su2024adversarial, rifatadversarial}: the attacker has access to the architecture and parameters of the initial deployed model $f_{\theta_0}$ (e.g., a publicly released checkpoint). The attacker may run $f_{\theta_0}$ locally to craft adversarial examples set ${\hat{X}}^{(t)}_A := {{X}}^{(t)}_A + {\delta}^{(t)}_{N_A}$, where ${\delta}^{(t)}_{N_A}$ is a set of perturbations, each defined in $\mathcal{X}$, and constrained within the $\ell_\infty$ budget $\epsilon$, i.e., $||{\delta}||_\infty \leq \epsilon$. However, the attacker does not know the deployed TTA algorithm $\mathcal{A}$, its hyperparameters, the sequence of adapted parameters $\{\theta_t\}_{t \ge 1}$, or any specific inputs from other users ${X}^{(t)}_B \cup {X}^{(t)}_V$.

%% file: 4_method.tex
\vspace{-0.14cm}
\section{Attack Design}
\vspace{-0.14cm}

\begin{figure*}[t]
    \centering
    \includegraphics[width=0.98\textwidth]{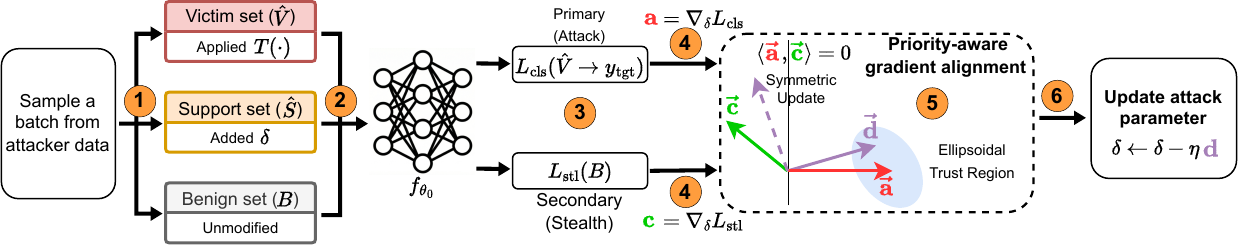}
    \caption{Overview of the attack generation using the proposed meta-learning workflow, where misalignment between attack and stealth gradients is resolved via priority-aware alignment.}
    \vspace{-1.2em}
    \label{fig:overall_r}
\end{figure*}

An overview of our attack generation pipeline is shown in Figure~\ref{fig:overall_r}. We simulate deployment-time TTA by repeatedly sampling batch from attacker's available data to emulate victim inputs (by applying trigger $T(\cdot)$), support inputs (by adding learnable perturbation $\delta$), and benign inputs. The attack is optimized in a meta-learning manner by jointly optimizing the attack objective and a newly introduced stealth objective (Section~\ref{sec:meta_attack}). Crucially, to alleviate the strong disagreement between objectives, we avoid symmetric combination and instead introduce a novel priority-aware gradient alignment that explicitly preserves the attack direction while maintaining stealth, with theoretical descent analysis (Section~\ref{sec:align_grad}). The resulting aligned gradient is used to update the attack parameters for the next meta-optimization step. After optimization, the learned perturbations are applied to the attacker’s inputs and used to deploy the attack.
\vspace{-0.1cm}
\subsection{Attack Generation via Meta-Learning} \label{sec:meta_attack}
\vspace{-0.1cm}
Under the threat model in Section~\ref{sec:threat_model}, we now detail how the attacker constructs concrete attack inputs under the grey-box setting. Since the attacker has access only to a local surrogate model $f_{\theta_0}$ and attacker-controlled data, the perturbations must generalize beyond attacker data. To this end, we introduce a meta-learning-based attack generation framework to learn the perturbations. Specifically, we parameterize the attack by a set of perturbations ${\delta}_{N_A}$, omitting the superscript $t$ for notational simplicity without loss of generality. We then construct multiple \emph{tasks}, as described below. The same ${\delta}_{N_A}$ is shared across many such tasks and will be optimized to work robustly across them.
\vspace{-0.8em}
\paragraph{Task construction.}
Each optimization iteration samples a batch from attacker data ${X}_A$ by randomly partitioning ${X}_A$ into three disjoint roles: \textbf{\emph{(i)}} a victim set ${\hat{V}}$, to which the trigger $T(\cdot)$ is applied and which should be classified as the target label ${y}_{\mathrm{tgt}}$; \textbf{\emph{(ii)}} a support set ${\hat{S}}$, which contains $N_S$ $(< N_A)$ samples carrying the corresponding learnable perturbation; and \textbf{\emph{(iii)}} a benign set ${B}$, which remains unmodified and is used to preserve stealth behavior. To improve generalization, the mixture ratios of the three sets are randomly sampled for each task from a uniform distribution. Across many randomly sampled tasks, all attacker samples are selected as support samples at different iterations, ensuring that the perturbations are optimized for the entire $X_A$.
\vspace{-0.8em}
\paragraph{Objectives.}
For each task, we define an attack loss that pushes triggered victim samples in ${\hat{V}}$ toward ${y}_{\mathrm{tgt}}$. This is implemented via the standard cross-entropy loss $L_\mathrm{cls}=\mathrm{CE}\big(f_{\theta_0}([{\hat{V}},{\hat{S}},{B}])\big|_{\hat V}, {y}_{\mathrm{tgt}}\big)$, where $\big|_{\hat V}$ restricts the batch outputs to the victim part only. This objective works via cross-sample coupling within a batch (see Appendix~\ref{appendix:chain}).
In addition, the attacker must preserve distributional stealth on non-victim inputs. Since TTA typically operates in a batch-wise mode, modifications in ${\hat{V}}$ and ${\hat{S}}$ inevitably alter the shared batch statistics used for $B$. To address this problem, we encourage prediction consistency between clean and perturbed versions of the same samples. Specifically, we propose a new distributional stealth loss that penalizes deviations in the model's output on benign inputs $B$ caused by the presence of $\hat{V}$ and $\hat{S}$ in the same batch. Formally,
\[
L_\mathrm{stl}=\mathrm{KL}\big(p_{\theta_0}([{V},{S},{B}])\big|_{B} \;||\; p_{\theta_0}([{\hat{V}},{\hat{S}},{B}])\big|_{B}\big),
\]
where $p_{\theta_0}(\cdot)=\mathrm{softmax}\left(f_{\theta_0}(\cdot)\right)$, KL is the Kullback-Leibler divergence, ${V}$ denotes the victim inputs without the trigger $T(\cdot)$, ${S}$ denotes the support inputs without the perturbations, and $\big|_{B}$ restricts the batch outputs to the benign part only. $L_\mathrm{stl}$ ensures that the benign predictions remain stable despite the shift in batch statistics caused by the attack.
\vspace{-0.8em}
\paragraph{Meta-learning update.}
Within one iteration, gradients from a predefined number of batches are aggregated to obtain batch-level estimates for the attack and stealth objectives. Based on these gradients, we construct an aligned gradient update direction ${d}^k$ (Section~\ref{sec:align_grad}), and update the perturbations via an $\ell_\infty$-constrained projected gradient step:
\begin{equation}
\label{eq:delta_update}
{\delta}^{k+1}_{N_A} = \Pi_{\mathcal{B}_\infty(\epsilon)} \bigl({\delta}^{k}_{N_A} - \eta d^k\bigr),
\end{equation}
where $k$ is the current step of the meta learning, $\eta>0$ is the step size, $\mathcal{B}_\infty(\epsilon)$ is the $\ell_\infty$ ball of radius $\epsilon$, and $\Pi_{\mathcal{B}_\infty(\epsilon)}$ denotes element-wise projection onto $\mathcal{B}_\infty(\epsilon)$. Repeating this procedure over multiple iterations yields perturbations that have been optimized across many randomly constructed tasks, i.e., in a meta-learning fashion. At deployment, the attacker applies the learned $\delta_{N_A}$ to its own queries $X_A$ to create $\hat{X}_A := X_A + \delta_{N_A}$ before sending them to the TTA service.
\vspace{-0.1cm}
\subsection{Attack Success and Stealth Misalignment} \label{sec:align_grad}
\vspace{-0.1cm}

\begin{wrapfigure}{r}{0.5\linewidth}
     \begin{subfigure}{.49\linewidth}
     \includegraphics[width=\textwidth]{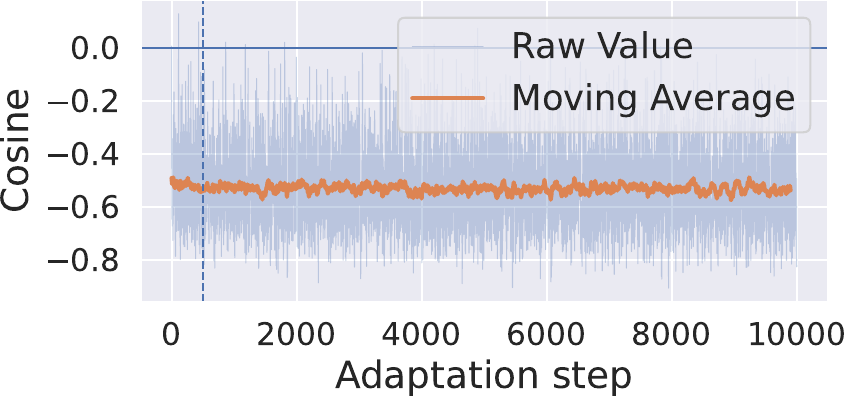}
     \caption{Cosine across steps.}
     \label{fig:cosine}
     \end{subfigure}%
     \hfill
    \begin{subfigure}{.49\linewidth}
     \includegraphics[width = \textwidth]{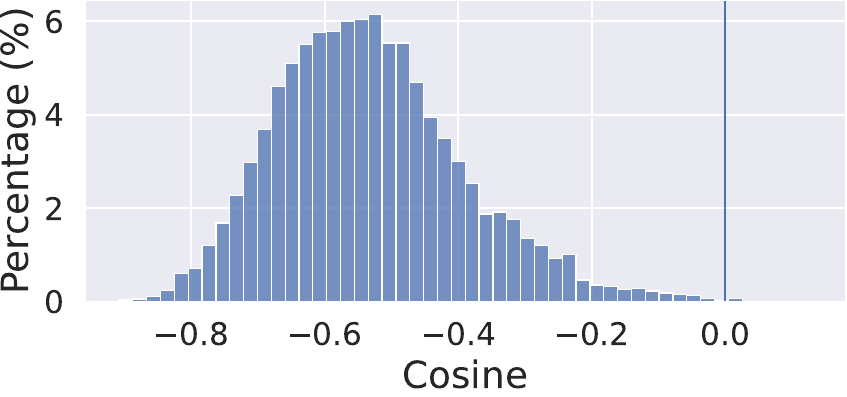}
     \caption{Frequency of cosine.}
     \label{fig:freq}
    \end{subfigure}
    \vspace{-0.5em}
    \caption{(a) Cosine similarity between gradients of $L_\mathrm{cls}$ and $L_\mathrm{stl}$ across attack optimization steps, where the vertical dashed line marks the 500 steps used in the experiments. (b) Frequency distribution of cosine similarity values. Nearly all steps ($99.9\%$) exhibit negative cosine similarity, indicating persistent gradient misalignment.}
    \vspace{-2em}
  \label{fig:cosine_a_c}
\end{wrapfigure}

In this section, we describe how the aligned direction $d^k$ is constructed. For brevity, we omit the superscript $k$. Empirically, we observe that the gradient of $L_\mathrm{cls}$ and the gradient of $L_\mathrm{stl}$ with respect to the perturbation are strongly antagonistic. As shown in Figure~\ref{fig:cosine_a_c}, the cosine similarity between the two gradients is negative in nearly all iterations ($99.9\%$), placing our optimization in the conflicting-gradients regime \cite{yu2020gradient}. 
Because a failed attack has zero utility regardless of its stealthiness, standard symmetric multi-task learning \cite{sener2018multi,liu2021conflict}, which treats all objectives equally, is unsuitable. Instead, we propose a priority-asymmetric objective where attack success is primary, and stealth is a secondary objective. Our update rule aligns closely with the attack gradient while penalizing movement toward stealth gradient that deviates from it (Section~\ref{sec:ellip_formulation}), which theoretically guarantees descent for the unprojected update even under severe gradient disagreement (Section~\ref{sec:theory}).

\vspace{-0.1cm}
\subsubsection{Priority-aware Gradient Alignment} \label{sec:ellip_formulation}
\vspace{-0.1cm}
Let $a := \nabla_{\delta} L_{\mathrm{cls}}(\delta_{N_A})$, $c := \nabla_{\delta} L_{\mathrm{stl}}(\delta_{N_A})$ denote the gradient of $L_\mathrm{cls}$ and $L_\mathrm{stl}$, respectively. To encode our attack-first preference, we formulate the update as a trust-region problem centered at the attack gradient $a$ under an ellipsoidal norm. Concretely, we seek an update direction $d$ that \textbf{\emph{(i)}} stays close to $a$, ensuring the update remains attack-driven, and \textbf{\emph{(ii)}} incorporates $c$ while penalizing movement toward directions that increasingly deviate from $a$. 

To encode \textbf{\emph{(i)}}, we additionally require that $d$ stays in a small neighbourhood of $a$. We measure distance using a Mahalanobis norm $\|v\|_M := \sqrt{v^\top M v}$ ($M \succ 0$),
and enforce a trust region of the form $\|d-a\|_M \le \rho$. This defines an ellipsoid centred at $a$ with radius $\rho>0$ in the $M$-metric, so any admissible $d$ remains close to $a$. 

To encode \textbf{\emph{(ii)}}, we choose the metric $M$ so that movement toward stealth directions that deviate from $a$ is penalized more strongly. Specifically, we set $M = I + \lambda u u^\top$, where $u$ denotes the deviation from $a$ toward $c$, and $\lambda \ge 0$ increases with the disagreement between $a$ and $c$. Intuitively, this choice leaves all directions orthogonal to $u$ unchanged, but makes movement along the deviation direction $u$ (i.e., deviating from $a$) more expensive: along $u$, the admissible Euclidean radius shrinks from $\rho$ to $\rho/\sqrt{1+\lambda}$. A precise construction of $u$ and $\lambda$ is given in Appendix~\ref{appendix:metric_details}.

We consider convex combinations of the attack and stealth gradients, i.e., $g_w = wa + (1-w)c$ for $w \in [0,1]$, which interpolates between purely attack-driven ($w=1$) and stealth-driven ($w=0$) directions. For an update $-\eta d$ (Eq.~\eqref{eq:delta_update}), the first-order approximation
$L(\delta-\eta d)\approx L(\delta)-\eta\nabla L^\top d$
shows that a larger inner product $\nabla L^\top d$ corresponds to a greater predicted loss reduction. However, a fixed weighting can mask a decline in one objective through improvement in the other. We therefore choose $d$ to maximize the worst-case first-order improvement over all $g_w$, while prioritizing the attack objective through the trust region centered at $a$. Setting the trust-region radius $\rho := \gamma \|a\|_2$ $(\gamma>0)$, gives the primal problem:
\begin{equation}\label{eq:primal}
\max_{d \in \mathbb{R}^d} \min_{w \in [0,1]} g_w^\top d
\quad \text{s.t.} \quad \|d-a\|_M \le \rho.
\end{equation}

We solve Eq.~\eqref{eq:primal} using Lagrangian \cite{boyd2004convex}, detailed in Appendix~\ref{appendix:detail_solve}. For a fixed $w$, the optimal update is:
\begin{equation}\label{eq:d_star}
d^\star(w) = a + \rho \frac{M^{-1} g_w}{\|g_w\|_{M^{-1}}},
\end{equation}
where $\|g_w\|_{M^{-1}} := \sqrt{g_w^\top M^{-1} g_w}$. We then have the following one-dimensional convex program:
\begin{equation}
\label{eq:final_min}
\min_{w \in [0,1]} a^\top g_w + \rho \|g_w\|_{M^{-1}}.
\end{equation}
We solve Eq.~\eqref{eq:final_min} over $w\in[0,1]$ using a scalar convex optimizer (e.g., Brent's method), obtain $w^\star$, and finally set $d^\star = d^\star(w^\star)$ via Eq.~\eqref{eq:d_star}.

\vspace{-0.1cm}
\subsubsection{Theoretical Understanding} \label{sec:theory}
\vspace{-0.1cm}
In this section, we characterize the effect of one update step on the perturbation $\delta_{N_A}$ when using the ellipsoidal direction $d^\star$ from Eq.~\eqref{eq:d_star}. 
We assume $L_{\mathrm{cls}}$ is $L$-smooth in $\delta_{N_A}$. The following theorem summarizes the behavior of one step along $d^\star$, with the proof deferred to Appendix~\ref{appendix:convergence_details}.

\begin{theorem}[Alignment-adaptive descent]
\label{thm:ellip_decrease_main}
Let $L_{\mathrm{cls}}$ be $L$-smooth, and let $d^\star$ be the ellipsoidal update direction with radius $\rho = \gamma \|a\|_2$ for $\gamma \in (0,1)$. Then there exists a factor $\xi \in \Big[\tfrac{1}{\sqrt{1+\lambda}},\,1\Big]$ that depends on how strongly $d^\star$ points into the deviation direction $u$ such that, for the step size $\eta := \frac{1-\gamma\xi}{L(1+\gamma\xi)^2}$, the update $\delta_{N_A}^{+} = \delta_{N_A} - \eta\, d^\star$ satisfies
\begin{equation}
\label{eq:ellip_decrease_main}
L_{\mathrm{cls}}(\delta_{N_A}^{+}) \le L_{\mathrm{cls}}(\delta_{N_A}) - \frac{(1-\gamma\xi)^2}{2L(1+\gamma\xi)^2}\,\|a\|_2^2.
\end{equation}
\end{theorem}

Theorem~\ref{thm:ellip_decrease_main} has two key implications.
\textbf{\emph{(i)}} In the isotropic case, $\lambda=0$ and $\xi=1$, recovering the standard Euclidean trust-region update ($M=I$).
\textbf{\emph{(ii)}} When $c$ deviates from $a$, $\lambda>0$ can yield $\xi<1$, resulting in a strictly stronger descent bound. Thus, the ellipsoidal geometry adapts to gradient disagreement by discouraging movement toward misaligned stealth directions, while reducing to the isotropic update when no disagreement is present.

Theorem~\ref{thm:ellip_decrease_main} analyzes the unprojected update, whereas our implementation projects the perturbation onto the feasible $\ell_\infty$ ball (Eq.~\ref{eq:delta_update}). In Appendix~\ref{appendix:projected_descent}, we derive a sufficient condition for descent under projection and empirically show that the projected update fails to decrease the attack objective in only a small fraction of iterations, with negligible impact on attack performance.

%% file: 5_experiment.tex
\vspace{-0.1cm}
\section{Experiment} \label{sec:exp}
\vspace{-0.1cm}

In this section, we first describe the experimental setup used to evaluate our proposed attack in Section~\ref{exp:setting}. We then present the main results in Section~\ref{exp:result}, followed by an ablation study in Section~\ref{exp:abl}. Finally, in Section~\ref{exp:defense}, we examine existing defenses' performance against our proposed attack. More experiments are presented in Appendix~\ref{appendix:more_exp}.
\vspace{-0.1cm}
\subsection{Experimental Setting} \label{exp:setting}
\vspace{-0.1cm}
\paragraph{Datasets and backbones.}
We evaluate the attack on standard benchmarks for TTA, namely CIFAR10-C, CIFAR100-C, and ImageNet-C. These datasets are constructed by applying a diverse set of synthetic corruptions to clean images \cite{hendrycks2019robustness}, thereby simulating realistic distribution shifts. Following common practice \cite{su2024adversarial, rifatadversarial}, for CIFAR10-C and CIFAR100-C, we adopt ResNet-32 as the backbone and evaluate performance across all 15 corruption types and all five severity levels; for ImageNet-C, we use ResNet-50 and report results at corruption severity level 3. We provide cross-dataset backbone results in Appendix~\ref{appendix:cross}.   
\vspace{-0.8em}
\paragraph{Baselines and TTA methods.} We compare our attack against two class-wise targeted TTA attacks, FCA \cite{rifatadversarial} and RTTDP \cite{su2024adversarial}, under the threat model presented in Section~\ref{sec:threat_model}. Following prior work \cite{wu2023uncovering, cong2024test, su2024adversarial, rifatadversarial}, we evaluate attacks under five BN-affine TTA methods: TENT \cite{wang2020tent}, RPL \cite{rusak2021if}, EATA \cite{niu2022efficient}, SAR \cite{niu2023towards}, and DeYO \cite{lee2024entropy}. More details on each TTA method are presented in Appendix~\ref{appendix:imp_detail}.
\vspace{-0.8em}
\paragraph{Evaluation protocol.}
We randomly select $5\%$ from distinct classes as triggered victims. We evaluate patch~\cite{gu2017badnets} and SIG~\cite{barni2019new} triggers (Figure~\ref{fig:trigger_vis}). Results with an imperceptible trigger are reported in Appendix~\ref{appendix:imperceptible_trigger}. We measure attack effectiveness by attack success rate (ASR) and stealth by entropy distribution drift (EDD) and augmentation disagreement inconsistency (ADI), where lower values indicate greater stealth. EDD captures shifts in the predictive-entropy distribution of non-targeted inputs, while ADI captures changes in their augmentation consistency relative to benign adaptation. We set $\epsilon=16/255$~\cite{su2024adversarial}. Further implementation details and sensitivity analyses are provided in Appendices~\ref{appendix:imp_detail} and~\ref{appendix:sen}, respectively.

\vspace{-0.1cm}
\subsection{Main Results} \label{exp:result}
\vspace{-0.1cm}

\input{main_sig}
\begin{figure}[ht]
    \centering
    \includegraphics[width=0.98\linewidth]{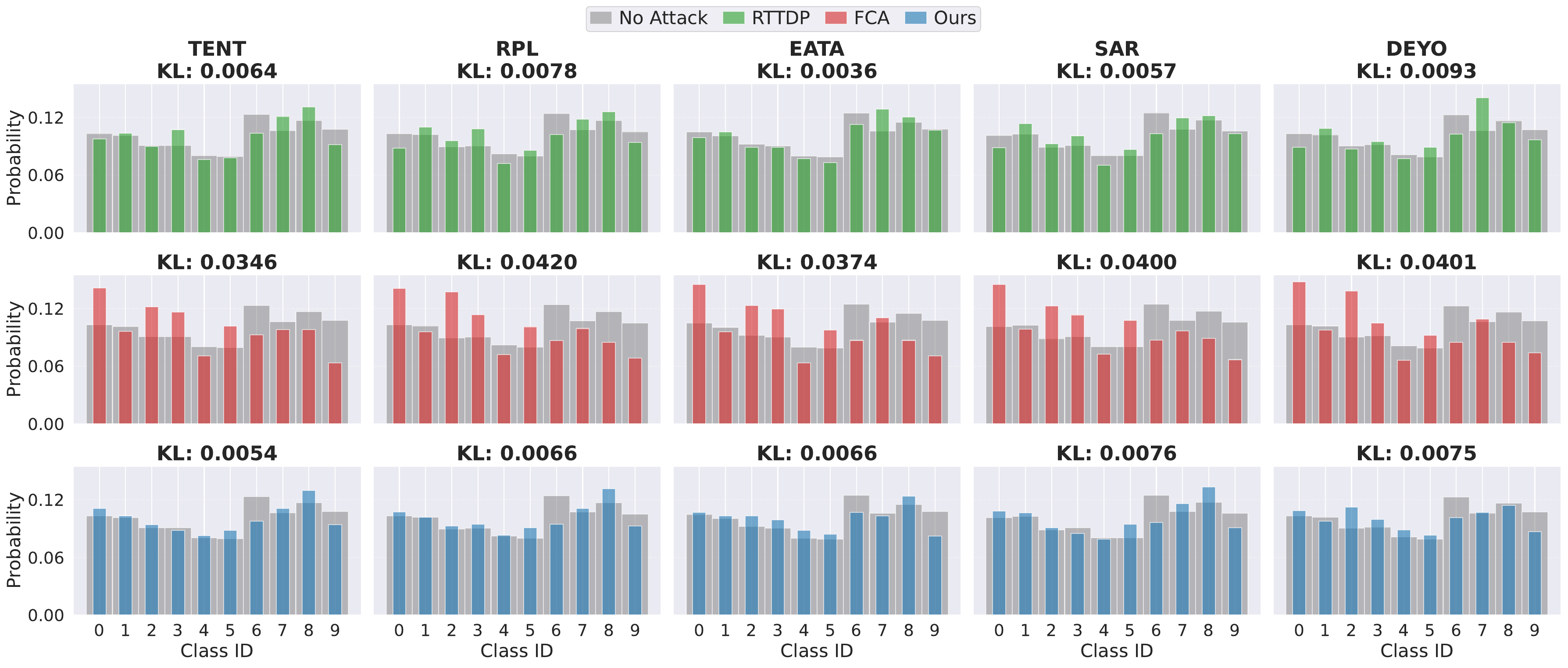}
    \vspace{-0.4em}
    \caption{Comparison of label distribution shifts induced by RTTDP, FCA, and ours on CIFAR-10-C.}
    \label{fig:main_stealth_10}
    \vspace{-1.2em}
\end{figure}

Table~\ref{tab:main_sig} compares our attack with FCA~\cite{rifatadversarial} and RTTDP~\cite{su2024adversarial}, both adapted to our threat model. Our method consistently achieves the highest ASR across all datasets, triggers, and TTA methods, frequently exceeding $80\%$ on CIFAR and $95\%$ on ImageNet-C. Adapting RTTDP and FCA to our meta-learning framework substantially improves their performance over their original formulations (values in parentheses), confirming that explicitly optimizing for the sample-wise threat model is important. Nevertheless, both baselines remain considerably weaker than our method, indicating that the proposed attack objective and priority-aware gradient alignment are critical for achieving high attack success under this setting.

Regarding stealthiness, across all three datasets, our method consistently achieves lower mean EDD and mean ADI than FCA while simultaneously obtaining substantially higher ASR. RTTDP achieves the lowest EDD and ADI, but this comes at the cost of much weaker attack effectiveness. These results indicate that our method achieves a favorable balance between attack success and label-free stealth, rather than improving ASR simply by inducing large changes in model behavior. Figure~\ref{fig:main_stealth_10} provides complementary evidence using the output-label distribution and KL divergence between the attacked and benign distributions. Class-wise attacks such as FCA produce noticeable probability spikes and larger distribution shifts (KL $>0.03$), whereas our trigger-defined attack remains close to the no-attack distribution (KL $= 0.006$), further demonstrating that high attack success can be achieved without creating an obvious concentration in the predicted target class.

\vspace{-0.1cm}
\subsection{Ablation Study} \label{exp:abl}
\vspace{-0.1cm}

Table~\ref{tab:ablation_study} analyzes the contribution of each component in our framework. Optimizing $L_{\mathrm{cls}}$ alone already yields a strong ASR of $87.43\%$, but produces clear probability spikes and substantially larger label-distribution shifts than our method (Appendix~\ref{appendix:label_dist}). Naively combining the attack and stealth objectives ($L_{\mathrm{cls}}+L_{\mathrm{stl}}$) achieves the lowest EDD and ADI, but reduces ASR to only $37.65\%$. This sharp trade-off is consistent with the severe gradient misalignment observed in Figure~\ref{fig:cosine_a_c}, and shows that simply emphasizing stealth can render the attack ineffective. Standard conflict-resolution methods do not adequately resolve this asymmetry. PCGrad~\cite{yu2020gradient} and CAGrad~\cite{liu2021conflict} achieve ASRs of only $38.55\%$ and $48.66\%$, respectively, as their symmetric treatment of the two objectives continues to sacrifice the primary attack objective. The Euclidean trust region (isotropic approach) better preserves attack effectiveness, reaching $79.88\%$ ASR, but still falls substantially below the attack-only baseline. In contrast, our priority-aware ellipsoidal alignment achieves the highest ASR of $91.03\%$, exceeding even $L_{\mathrm{cls}}$ alone and outperforming the isotropic trust region by more than $11$ percentage points. While its EDD and ADI remain comparable to the attack-only setting, the label-distribution analysis in Appendix~\ref{appendix:label_dist} shows substantially improved distributional behavior. Overall, the ablation confirms the motivation of our formulation: rather than seeking a symmetric compromise that substantially weakens the attack, our method preserves the primary attack objective while incorporating stealth as a secondary objective.

\vspace{-0.1cm}
\subsection{Existing Defenses Performance} \label{exp:defense}
\vspace{-0.1cm}

\begin{figure}[t]
\begin{minipage}[t]{0.62\linewidth}
    \vspace{-6.3em}
    \small
    \centering
    \captionof{table}{ASR, EDD, and ADI of attacks averaged across various TTA methods, SIG trigger, CIFAR-10-C dataset.}
    \label{tab:ablation_study}
    \begin{tabular}{lccc}
    \toprule
    Method & ASR & EDD & ADI \\
    \midrule
    $L_{\mathrm{cls}}$                   & 87.43 & 0.245 & 20.12 \\
    $L_{\mathrm{cls}}+L_{\mathrm{stl}}$  & 37.65 & \textbf{0.078} & \textbf{5.81} \\
    W/ PCGrad                            & 38.55 & 0.153 & 10.17 \\
    W/ CAGrad                            & 48.66 & 0.165 & 10.96 \\
    W/ Euclidean trust region            & 79.88 & 0.218 & 16.64 \\
    Ours                                 & \textbf{91.03} & 0.225 & 20.19 \\
    \bottomrule
    \end{tabular}

    \scriptsize
    \captionof{table}{ASR of our attack under CoTTA defense~\cite{wang2022continual}.}
    \label{tab:cotta}
    \begin{tabular}{l c c c}
    \toprule
    {\textbf{Trigger}} & \textbf{No attack} & \textbf{Ours} & \textbf{Ours w/ Aug} \\
    \midrule
    SIG
    & 3.14
    & 13.60
    & \textbf{41.44} \\
    Patch
    & 2.48
    & 12.95
    & \textbf{41.59} \\
    \bottomrule
    \end{tabular}
    
\end{minipage}
\hfill
\begin{minipage}[t]{0.35\linewidth}
    \includegraphics[width=0.96\linewidth]{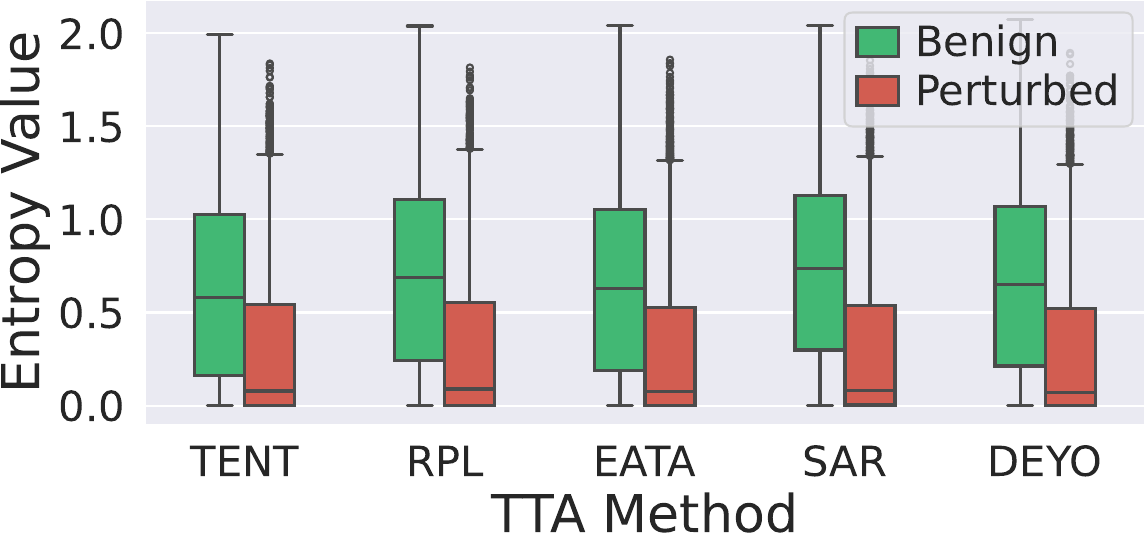}
    \vspace{-0.1cm}
    \captionof{figure}{Entropy of benign and perturbed samples.}
    \label{fig:entropy}
    \includegraphics[width=0.96\linewidth]{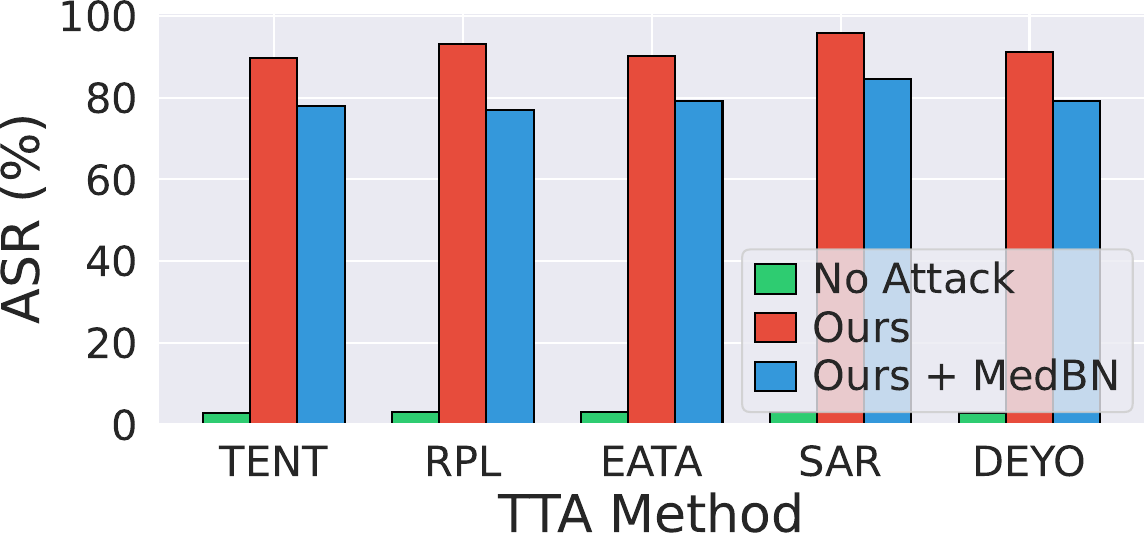}
    \vspace{-0.1cm}
    \captionof{figure}{ASR of our attack under MedBN defense~\cite{park2024medbn}.}
    \label{fig:medbn}
\end{minipage}
\vspace{-1.2em}
\end{figure}

\paragraph{Sample filtering.}
A potential defense against our attack is sample filtering, which discards high-entropy samples to prevent noisy data from degrading TTA performance~\cite{niu2022efficient, niu2023towards, gong2023sotta}. 
However, as illustrated in Figure~\ref{fig:entropy}, our attack generates samples with entropy values even lower than those of benign samples, as the attack optimization minimizes uncertainty to ensure target success. 
Thus, entropy-based filtering fails to remove our adversarial inputs. 
This ineffectiveness is empirically confirmed by the high ASR achieved against EATA~\cite{niu2022efficient} and SAR~\cite{niu2023towards} in Table~\ref{tab:main_sig}.

\vspace{-0.8em}
\paragraph{TTA-specific defense.} Figure~\ref{fig:medbn} reports performance against MedBN defense~\cite{park2024medbn}, which replaces standard BN with Median BN for robust batch statistics. Our attack still achieves high ASR ($\approx80\%$). MedBN primarily hinders optimization through gradient masking: the derivative of the median is zero for non-median samples, obstructing gradient flow. However, under our grey-box threat model, the attacker crafts perturbations using the original pre-deployment model equipped with standard BN, which provides a smooth gradient landscape for optimization. Since shifts in batch statistics induced under BN can transfer to the median-based statistics, the resulting perturbations remain effective against MedBN at test time.
\vspace{-0.8em}
\paragraph{Exponential moving average (EMA).} Another potential defense involves using an EMA of the model weights during adaptation, rather than instantaneous updates \cite{wang2022continual}. This strategy initially mitigates our attack, reducing the ASR to about 13\% (Table~\ref{tab:cotta}). However, by simply applying data augmentations (e.g., blur, crop, flip) to the adversarial samples, we can recover the ASR to roughly 41\%. This occurs because augmentation forces the generation of robust and consistent gradients that accumulate in the EMA buffer, rather than brittle noise that gets averaged out. Notably, this recovery is achieved without explicitly optimizing the attack against the EMA, suggesting that simple heuristics are sufficient to compromise this defense.

%% file: main_sig.tex
\begin{table*}[ht]
\centering
\caption{ASR, Mean EDD, and Mean ADI of attacks under various TTA methods. The best results are highlighted in \textbf{bold}. $\star$ denotes methods adapted to our threat model (see Appendix~\ref{appendix:imp_detail} for details), and values in parentheses report the original results without adaptation.}
\label{tab:main_sig}
\resizebox{\linewidth}{!}{
\begin{tabular}{cc|ccccc|ccccc|cc|}
\toprule
\multirow{2}{*}{\textbf{Dataset}} & \multirow{2}{*}{{\textbf{Attack}}} & \multicolumn{5}{c|}{\textbf{SIG Trigger}} & \multicolumn{5}{c|}{\textbf{Patch Trigger}} & \multirow{2}{*}{\textbf{\begin{tabular}[c]{@{}c@{}}Mean\\EDD\end{tabular}}} & \multirow{2}{*}{\textbf{\begin{tabular}[c]{@{}c@{}}Mean\\ADI\end{tabular}}}  \\
\cmidrule(lr){3-12}
& & {{TENT}} & {RPL} & {EATA} & {SAR} & {DeYO} & {{TENT}} & {RPL} & {EATA} & {SAR} & {DeYO} &  \\
\midrule
\multirow{4}{*}{\textbf{\begin{tabular}[c]{@{}c@{}}CIFAR-10-C \\ (ResNet32)\end{tabular}}} 
& No attack  &  2.85 & 3.07 &  3.10 & 3.04 &  2.85 & 2.77 & 3.16 & 3.02 & 3.64&2.97 & - & - \\
& RTTDP$\star$  &25.39 (7.56)  &29.65 (8.17) &22.80 (7.53)  &37.70 (8.30)  &26.10 (7.51) &32.72 (6.55)&42.54 (7.57)&35.98 (6.57)&60.58 (8.95)&35.47 (6.76) & \textbf{0.16} & \textbf{18.08} \\
& FCA$\star$  & 54.80 (6.67) & 59.62 (7.27)  &51.99 (6.76) &63.26 (7.29)  &54.09 (6.52) &50.77 (5.43)&58.36 (6.40)&50.15 (5.68)&58.97 (7.45) &51.74 (5.74) & 0.60 & 28.06 \\
&\cellcolor{hl} Ours  &\cellcolor{hl}\textbf{89.59}     &\cellcolor{hl}\textbf{93.17}  &\cellcolor{hl}\textbf{90.21}  &\cellcolor{hl}\textbf{95.90} &\cellcolor{hl}\textbf{91.15} &\cellcolor{hl}\textbf{79.28}
&\cellcolor{hl}\textbf{86.78}
&\cellcolor{hl}\textbf{80.34}
&\cellcolor{hl}\textbf{93.33}
&\cellcolor{hl}\textbf{82.48} &\cellcolor{hl}{0.23} &\cellcolor{hl}{21.14}  \\
\midrule
\multirow{4}{*}{\textbf{\begin{tabular}[c]{@{}c@{}}CIFAR-100-C \\ (ResNet32)\end{tabular}}}  
& No attack  & 0.07   & 0.09  & 0.11  & 0.09  & 0.09 &0.87&0.91&0.96&0.91&0.82 & - & - \\
& RTTDP$\star$ &30.17 (0.15)  &39.31 (0.12) &37.35 (0.14)  &41.92 (0.15)  &31.11 (0.13)                &64.80 (0.40)&76.16 (0.40)&75.30 (0.39)&77.84 (0.39)&69.44 (0.35) & \textbf{0.10} & \textbf{14.39} \\
& FCA$\star$  &45.81 (0.15)  &49.13 (0.13)  &51.22 (0.14)  &54.49 (0.15)  &43.65 (0.14)           &73.03 (0.40)&76.88 (0.40)&80.61 (0.40)&80.84 (0.40)& 70.95 (0.36) & 0.62 & 19.09  \\
&\cellcolor{hl} Ours  &\cellcolor{hl}\textbf{66.93}     &\cellcolor{hl}\textbf{76.17}  &\cellcolor{hl}\textbf{76.42}  &\cellcolor{hl}\textbf{81.49}  &\cellcolor{hl}\textbf{70.66} &\cellcolor{hl}\textbf{88.04}
&\cellcolor{hl}\textbf{91.82}
&\cellcolor{hl}\textbf{91.83}
&\cellcolor{hl}\textbf{93.47}
&\cellcolor{hl}\textbf{88.97} &\cellcolor{hl}{0.16} &\cellcolor{hl}{15.36}     \\
\midrule
\multirow{4}{*}{\textbf{\begin{tabular}[c]{@{}c@{}}ImageNet-C \\ (ResNet50)\end{tabular}}}  
& No attack  & 0.00  & 0.00  & 0.09  & 0.00  & 0.09 & 0.11  & 0.11  & 0.05  & 0.11  & 0.05 & - & -  \\
& RTTDP$\star$  &49.50 (0.00)  &54.08 (0.00)  &37.14 (0.00)  &53.54 (0.00)  &41.67 (0.00)      &32.67 (0.00)  &31.63 (0.00)  &20.95 (0.00)  &32.32 (0.00)  &21.30 (0.00) & \textbf{0.15} & \textbf{17.95}  \\
& FCA$\star$  &92.08 (0.00) &94.90 (0.00)  &86.67 (0.00)  &95.96 (0.00)  &87.04 (0.00)         &84.16 (0.00)  &85.71 (0.00)  &73.33 (0.00)  & 81.82 (0.00)  &71.96 (0.00)  & 0.60 & 31.07   \\
&\cellcolor{hl} Ours &\cellcolor{hl}\textbf{96.79}  &\cellcolor{hl}\textbf{96.99} &\cellcolor{hl}\textbf{95.22} &\cellcolor{hl}\textbf{96.98}  &\cellcolor{hl}\textbf{95.49} &\cellcolor{hl}\textbf{92.29}
&\cellcolor{hl}\textbf{92.49}
&\cellcolor{hl}\textbf{88.48}
&\cellcolor{hl}\textbf{92.28}
&\cellcolor{hl}\textbf{89.21}  &\cellcolor{hl}{0.24}  &\cellcolor{hl}{22.29}    \\
\bottomrule

\end{tabular}
}
\vspace{-1em}
\end{table*}

%% file: 7_conclusion.tex
\vspace{-0.1cm}
\section{Conclusion}
\vspace{-0.1cm}
In this paper, we formalize a sample-wise targeted threat model for TTA, where only trigger-carrying inputs are misclassified while non-triggered inputs maintain benign-like prediction behavior. We propose a meta-learning framework enhanced by a novel priority-aware gradient alignment, which mitigates the misalignment between attack success and stealth, supported by theoretical descent analysis of the attack objective. Our approach significantly outperforms existing baselines, achieving high targeted attack success rates while remaining consistent with benign adaptation under multiple label-free stealth metrics. Additionally, we investigate the attack effectiveness against existing defenses, highlighting security risks in deployment-time adaptation.

%% file: 8_appendix.tex
\newpage
\appendix

\section{Extended Related Works} \label{appendix:more_rw}

\paragraph{Test-time Adaptation.}
TTA updates model parameters at inference time using unlabeled test data to handle distribution shift \cite{liang2025comprehensive}. Unlike training-time domain adaptation \cite{li2024comprehensive}, it assumes no access to source data or labels at deployment. TTA methods are commonly grouped by the adaptation signal: (i) \emph{entropy/confidence-based} updates that minimize prediction entropy \cite{wang2020tent}, improved via sample filtering \cite{lee2023towards,lee2024stationary,lee2024entropy}, conservative entropy minimization \cite{zhang2024come,han2025ranked}, and confidence regularization \cite{hu2025beyond}; and (ii) \emph{self-training/consistency-based} updates using sharpness-aware adaptation \cite{niu2023towards}, robustness-regularized self-training \cite{rusak2021if,jang2022test,ma2024improved}, or EMA teacher-student guidance \cite{wang2022continual,lyu2024variational,tian2024parameter}. Since TTA relies on online updates from the unlabeled test stream, implicitly trusting the test stream, model behavior becomes dependent on the adaptation trajectory, thereby exposing an attack surface. In this work, we study adversaries that exploit this by targeting the adaptation process.

\paragraph{Attacks on Test-time Adaptation.}
Since TTA updates model parameters from unlabeled test inputs, adversaries can manipulate a subset of the test samples to steer the adaptation trajectory. Existing attacks \cite{wu2023uncovering, cong2024test, su2024adversarial, rifatadversarial} mainly demonstrate untargeted degradation of post-adaptation performance. In practice, however, attackers often benefit more from targeted manipulation. \cite{wu2023uncovering} demonstrates targeted attacks in a strong setting (white-box access and access to victim data), and subsequent works relax these assumptions, including grey-box settings \cite{rifatadversarial} and realistic test-time data poisoning without requiring victim-data access \cite{su2024adversarial}. 
However, these attacks are \emph{class-wise targeted}, and thus become unstealthy as they unnaturally concentrate predictions on the target class (Figure~\ref{fig:example}). In contrast, we study a \emph{sample-wise targeted} threat model: only inputs carrying the attacker’s trigger should be redirected to the target label, while non-triggered inputs should maintain a label distribution consistent with the no-attack baseline (Table~\ref{tab:taxo}). 

\paragraph{Connection to Backdoor Attacks.} Our threat model adopts the concept of trigger-activated misbehavior from traditional backdoor attacks~\cite{gu2017badnets,saha2020hidden,li2022backdoor}, utilizing triggers that can be realized digitally (e.g., pixel patterns~\cite{gu2017badnets}, steganography~\cite{li2021invisible}) or physically (e.g., patches or accessories in the scene~\cite{wenger2021backdoor,eykholt2018robust}). However, standard backdoors require compromising the training data (poisoning) or the model weights (trojaning) prior to deployment, which are often infeasible. In contrast, our attack operates strictly at test-time, exploiting the online adaptation mechanism of TTA. This allows an adversary to induce backdoor-like behavior in a clean, pre-trained model on the fly, without ever having accessed the training phase.

\paragraph{Multi-Task Learning and Conflicting Gradients}
Multi-task learning (MTL) \cite{zhang2021survey} optimizes shared parameters for multiple objectives, but often suffers from gradient interference \cite{yu2020gradient}, where misaligned task gradients lead to unstable training or poor trade-offs. Prior work mitigates this via loss reweighting \cite{kendall2018multi,chen2018gradnorm} or multi-objective/gradient-balancing updates \cite{sener2018multi,yu2020gradient,liu2021conflict}. Our problem is closely related: the attack and stealth objectives may produce gradients that are partially misaligned, or even directly conflicting when their inner product is negative. However, unlike standard MTL formulations that balance tasks symmetrically or approximate Pareto-optimal trade-offs, our setting is priority-asymmetric: maximizing attack success is the primary goal, while maintaining benign-like behavior is a secondary objective. This motivates a priority-aware alignment strategy that keeps the update close to the attack gradient and increasingly penalizes movement toward stealth-gradient directions as their angular disagreement with the attack direction grows, rather than treating both objectives equally as in conventional MTL.

\section{Causal Link Through Shared BatchNorm Statistics} \label{appendix:chain}

Fundamentally, our attack exploits the cross-sample dependency in standard BatchNorm-affine TTA methods, consistent with observations in prior works~\cite{wu2023uncovering, cong2024test, su2024adversarial, rifatadversarial}. During a mixed-batch forward pass, the normalized representation of any victim sample is dynamically coupled to the support set through the shared batch mean and variance.

Concretely, let $h_i^c$ denote the pre-BN activation of sample $i$ in channel $c$, and let the batch of $m$ samples be formed by the concatenation $[\hat V,\hat S,B]$, representing the victim, support, and clean sets, respectively. For a BN layer,
\[
\mu_c=\frac{1}{m}\sum_{k=1}^m h_k^c,\qquad
(\sigma_c^2)=\frac{1}{m}\sum_{k=1}^m (h_k^c-\mu_c)^2,
\]
and the normalized activation of a victim sample $i\in \hat V$ is $z_i^c=(h_i^c-\mu_c)/\sqrt{\sigma_c^2+\epsilon}$. Although the attack loss $L_{\mathrm{cls}}$ is evaluated only on the triggered victim samples, its gradient with respect to a support perturbation $\delta_j$ on $j\in \hat S$ is non-zero because $\delta_j$ changes the shared BN statistics:
\[
\frac{\partial L_{\mathrm{cls}}}{\partial \delta_j}
=
\sum_{i\in \hat V,c}
\frac{\partial L_{\mathrm{cls}}}{\partial z_i^c}
\left(
\frac{\partial z_i^c}{\partial \mu_c}\frac{\partial \mu_c}{\partial h_j^c}
+
\frac{\partial z_i^c}{\partial \sigma_c^2}\frac{\partial \sigma_c^2}{\partial h_j^c}
\right)
\frac{\partial h_j^c}{\partial \delta_j}.
\]

Through this cross-sample gradient flow, the support perturbations are learned to steer the shared BN normalization in a direction that makes the triggered victim samples more likely to be classified as the target label.

\section{Misalignment Metric Construction}\label{appendix:metric_details} 
We instantiate $M$ to penalize movement toward directions that deviate from the primary attack gradient $a$. For $\|a\|_2>0$ and $\|c\|_2>0$, let
\[
\hat a := \frac{a}{\|a\|_2},
\qquad
\hat c := \frac{c}{\|c\|_2},
\qquad
s := \hat a^\top \hat c,
\]
If either gradient has zero norm, we fall back to the attack-only direction.
Here, $s$ is the cosine similarity between the attack gradient and the stealth gradient. A larger angular disagreement between $a$ and $c$ indicates that the stealth gradient is less aligned with the primary attack objective.

We define the deviation axis as the direction from the attack gradient toward the stealth gradient:
\[
u :=
\begin{cases}
(\hat c-\hat a)/\|\hat c-\hat a\|_2, 
& \text{if } \|\hat c-\hat a\|_2 > \varepsilon,\\[0.3em]
0, & \text{otherwise},
\end{cases}
\]
for a small numerical threshold $\varepsilon>0$. 
Next, we set the anisotropy strength $\lambda\ge 0$ as a monotone function of the angular disagreement between $a$ and $c$:
\[
\lambda := \kappa (1-s),
\]
where $\kappa>0$ is a scalar hyperparameter. Thus, $\lambda$ is small when $c$ is aligned with $a$, increases as $c$ moves away from $a$, and is largest when $c$ points in the opposite direction of $a$.

The resulting metric is
\begin{equation}
\label{eq:metric_M}
M = I + \lambda\, u u^\top.
\end{equation}
Since $\lambda\ge 0$, this matrix is positive definite and satisfies $M\succeq I$.

To see its effect, consider a displacement purely along the deviation axis, i.e., $d - a = t u$,
\[
\|d-a\|_M^2 = \|t u\|_M^2 = (1+\lambda)\,\|t u\|_2^2 = (1+\lambda)\,t^2. \qquad (\text{since }\|u\|_2 = 1)
\]
Therefore,
\[
\|d-a\|_M=\|t u\|_M \;=\; \sqrt{1+\lambda}\,|t|.
\]
Under the trust-region constraint $\|d-a\|_M\le \rho$, this implies
\[
|t| \le \frac{\rho}{\sqrt{1+\lambda}}.
\]
By contrast, a Euclidean trust region $\|d-a\|_2\le \rho$ would allow $|t|\le \rho$ along the same direction. Hence, using $M$ shrinks the admissible movement along the deviation axis by a factor of $1/\sqrt{1+\lambda}$.

This gives a selective penalty: movement in the direction that deviates from the attack gradient toward the stealth gradient is suppressed, while movement orthogonal to $u$ receives no additional penalty beyond the Euclidean term.

\section{Solving Ellipsoidal Trust-region Problem}\label{appendix:detail_solve}
Setting the trust-region radius $\rho := \gamma \|a\|_2$ $(\gamma>0)$, we have the primal problem in Eq.~\eqref{eq:primal}:
\[
\max_{d \in \mathbb{R}^d} \min_{w \in [0,1]} g_w^\top d
\quad \text{s.t.} \quad \|d-a\|_M \le \rho.
\]

The objective is linear in $d$ and $g_w$ is affine in $w$, while the feasible set in $d$ is convex. Hence, the inner-max outer-min problem satisfies the conditions of Sion's minimax theorem \cite{sion1958general}, and we may swap the order of $\max$ and $\min$:
\begin{equation}\label{eq:primal_swap}
\min_{w \in [0,1]} \max_{d: \|d-a\|_M \le \rho} g_w^\top d.
\end{equation}
For a fixed $w$, the inner problem in Eq.~\eqref{eq:primal_swap} is a problem of maximizing a linear functional $g_w^\top d$ over an ellipsoid. Its solution can be obtained in closed form via the Lagrangian \cite{boyd2004convex}. Consider:
\[
\mathcal{L}(d,\mu) = g_w^\top d - \mu \big( (d-a)^\top M (d-a) - \rho^2 \big),
\qquad \mu \ge 0.
\]
Stationarity with respect to $d$ gives:
\[
\nabla_d \mathcal{L}
= g_w - 2\mu M (d-a) = 0
\quad\Rightarrow\quad
d = a + \frac{1}{2\mu} M^{-1} g_w.
\]
Since the objective is linear and the feasible set is bounded, the optimum lies on the boundary of the ellipsoid, so the constraint is active at $d^\star$, and we have
\[
\|d^\star - a\|_M^2
= \rho^2
\quad\Longleftrightarrow\quad
\left\|\frac{1}{2\mu^\star} M^{-1} g_w\right\|_M^2
= \rho^2.
\]
Using $\|v\|_M^2 = v^\top M v$ and the definition of the dual norm $\|g_w\|_{M^{-1}} := \sqrt{g_w^\top M^{-1} g_w}$, this yields
\[
\frac{1}{4(\mu^\star)^2} \, g_w^\top M^{-1} g_w
= \rho^2
\quad\Rightarrow\quad
\mu^\star
= \frac{1}{2\rho} \, \|g_w\|_{M^{-1}}.
\]
Substituting $\mu^\star$ back into the expression for $d$ gives the optimal update for fixed $w$:
\[
d^\star(w) = a + \rho \, \frac{M^{-1} g_w}{\|g_w\|_{M^{-1}}}.
\]

Note that if \(\|g_{w^\star}\|_{M^{-1}}\le \varepsilon\), we fall back to the attack-only direction \(d^\star=a\). This avoids the degenerate case where the normalized trust-region direction is undefined.

\section{Proof of Theorem~\ref{thm:ellip_decrease_main}}\label{appendix:convergence_details}
\paragraph{Unit $M$-norm.} For any $w\in[0,1]$, the solution of the inner trust-region problem is
\[
d^\star(w)=a+\rho p_w,
\qquad
p_w:=\frac{M^{-1}g_w}{\|g_w\|_{M^{-1}}}.
\]
Let $w^\star$ be a minimizer of Eq.~\eqref{eq:final_min}, and define
\[
p^\star:=p_{w^\star},
\qquad
d^\star:=d^\star(w^\star)=a+\rho p^\star .
\]

$p^\star$ has unit $M$-norm:
\[
\|p^\star\|_M^2
=
(p^\star)^\top M p^\star
=
\frac{g_{w^\star}^\top M^{-1} M M^{-1}g_{w^\star}}
     {g_{w^\star}^\top M^{-1}g_{w^\star}}
=
1.
\]

\paragraph{Bounds on $\xi$.} 
Define $\xi := \|p^\star\|_2$. Since $M\succeq I$, we have
\[
\xi=\|p^\star\|_2 \le \|p^\star\|_M=1.
\]

Since $M=I+\lambda uu^\top$ with $\|u\|_2=1$, for any vector $p$ we can write
\[
p_u := (u^\top p)u,
\qquad
p_{\mathrm{rest}} := p-p_u,
\qquad
u^\top p_{\mathrm{rest}}=0.
\]
Then
\[
\|p\|_M^2 = p^\top Mp = \|p_{\mathrm{rest}}\|_2^2+(1+\lambda)\|p_u\|_2^2.
\]
Applying this to $p=p^\star$ and using $\|p^\star\|_M=1$, we get
\begin{align*}
\|p^\star_{\mathrm{rest}}\|_2^2+(1+\lambda)\|p^\star_u\|_2^2
&\le (1+\lambda)\left(\|p^\star_{\mathrm{rest}}\|_2^2+\|p^\star_u\|_2^2\right), \\
\Longleftrightarrow \|p^\star\|_M^2 = 1 &\le (1+\lambda)\|p^\star\|_2^2, \\
\Longleftrightarrow \frac{1}{\sqrt{1+\lambda}} &\le \|p^\star\|_2, \\
\Longleftrightarrow \frac{1}{\sqrt{1+\lambda}} &\le \xi \le 1.
\end{align*}

\paragraph{Descent bound.} Using $d^\star=a+\rho p^\star$ and $\rho=\gamma\|a\|_2$, we obtain
\begin{align}
a^\top d^\star
&= a^\top a + \rho a^\top p^\star
   = \|a\|_2^2 + \rho a^\top p^\star
   \nonumber\\
&\ge \|a\|_2^2 - \rho \|a\|_2\|p^\star\|_2
   \qquad\text{(Cauchy-Schwarz and $-\|x\|\|y\|\le x^\top y$)}
   \nonumber\\
&= \|a\|_2^2 - \rho \|a\|_2\xi
   \qquad\text{($\|p^\star\|_2 = \xi$)}
   \nonumber\\
&= (1-\gamma\xi) \|a\|_2^2.
\label{eq:aTd_lower_bound}
\end{align}
Thus, when $a\neq 0$, $\xi\le 1$, and $\gamma<1$, $-d^\star$ is a descent direction for $L_{\mathrm{cls}}$.

\paragraph{Step norm bound.} We also have the following norm bound
\begin{equation}
\label{eq:d_norm_bound_appendix}
\|d^\star\|_2
=
\|a+\rho p^\star\|_2
\le
\|a\|_2+\rho\|p^\star\|_2
=
(1+\gamma\xi)\|a\|_2 .
\end{equation}

\paragraph{Applying smoothness.} Since $L_{\mathrm{cls}}$ is $L$-smooth, for any $\eta>0$, the descent lemma~\cite{nesterov2013introductory} gives
\begin{equation}
\label{eq:smoothness_appendix}
L_{\mathrm{cls}}(\delta_{N_A}-\eta d^\star)
\le
L_{\mathrm{cls}}(\delta_{N_A})
-\eta a^\top d^\star
+\frac{L\eta^2}{2}\|d^\star\|_2^2 .
\end{equation}
Substituting the bounds \eqref{eq:aTd_lower_bound} and \eqref{eq:d_norm_bound_appendix} into \eqref{eq:smoothness_appendix} gives
\[
L_{\mathrm{cls}}(\delta_{N_A}-\eta d^\star)
\le
L_{\mathrm{cls}}(\delta_{N_A})
-\eta(1-\gamma\xi)\|a\|_2^2
+\frac{L\eta^2}{2}(1+\gamma\xi)^2\|a\|_2^2 .
\]

Choose
\[
\eta = \frac{1-\gamma\xi}{L(1+\gamma\xi)^2}.
\]
Then
\begin{align*}
L_{\mathrm{cls}}(\delta_{N_A} - \eta d^\star)
&\le
L_{\mathrm{cls}}(\delta_{N_A})
\;-\;
\eta(1-\gamma\xi)\,\|a\|_2^2
\;+\;
\frac{\eta(1-\gamma\xi)}{2}\,\|a\|_2^2
\\
&=
L_{\mathrm{cls}}(\delta_{N_A})
\;-\;
\frac{\eta(1-\gamma\xi)}{2}\,\|a\|_2^2
\\
&=
L_{\mathrm{cls}}(\delta_{N_A})
\;-\;
\frac{(1-\gamma\xi)^2}{2L\,(1+\gamma\xi)^2}\,\|a\|_2^2.
\end{align*}
This completes the proof.
\qed

\section{Attack Implementation} \label{appendix:algo}

Algorithm~\ref{alg:attack} summarizes the meta-learning procedure used to generate the attack. Starting from an initial zero perturbations $\delta_{N_A}$, we run $K$ outer iterations. At each iteration, we sample $T$ batches from the attacker’s data $X_A$, where each batch includes three components as described in Section~\ref{sec:meta_attack}. For each batch, use the base model $f_{\theta_0}$ to evaluate the attack loss $L_{\mathrm{cls}}^{\tau}$ and the distributional stealth loss $L_{\mathrm{stl}}^{\tau}$ under the current perturbation. The corresponding gradients with respect to $\delta_{N_A}$ are accumulated and averaged across batches to obtain batch-level attack and stealth gradients $a$ and $c$. These aggregated gradients are then fed into the \textsc{EllipsoidalDirection}, which solves the ellipsoidal trust-region problem from Eqs.~\eqref{eq:primal}--\eqref{eq:final_min} and returns an aligned direction $d^k$ that prioritizes the attack objective while penalizing movement toward misaligned stealth directions. It is important to note that, although we write $M=I+\lambda uu^\top$, we do not instantiate the full $F\times F$ matrix, where $F$ is the flattened perturbation dimension. Since $M$ is a rank-one perturbation of the identity, we compute $M^{-1}g_w = g_w-\frac{\lambda}{1+\lambda}u(u^\top g_w)$, so each trust-region evaluation only costs $O(F)$ time and memory.
Finally, we update $\delta_{N_A}$ via an $\ell_\infty$-projected gradient step along $-d^k$, enforcing the perturbation budget $\|\delta_{N_A}\|_\infty \le \epsilon$. After $K$ iterations, the learned perturbations are added to all attacker inputs to produce the perturbed attacker data $\hat{X}_A = X_A + \delta_{N_A}$ used to steer the TTA process at deployment.

\input{algo}

\section{Descent Analysis under $\ell_\infty$ Projection} \label{appendix:projected_descent}

\paragraph{Sufficient decent condition.} Theorem~\ref{thm:ellip_decrease_main} analyzes the unprojected update $\delta^{+}=\delta-\eta d^\star$. In the actual attack, however, the perturbation is constrained to an $\ell_\infty$ ball, and the update is therefore $\delta^{+} = \Pi_{\mathcal{B}_\infty(\epsilon)} \left(\delta-\eta d^\star\right)$. We next establish a sufficient condition under which this projected update still decreases the attack objective.

Define the projected gradient mapping $G_\eta := \frac{\delta-\delta^{+}}{\eta}$. Since the current iterate $\delta$ is feasible, the optimality condition of Euclidean projection gives
\begin{equation} \label{eq:projection_optimality}
(d^\star)^\top G_\eta
\geq
\|G_\eta\|_2^2.
\end{equation}

Let $a := \nabla_{\delta} L_{\mathrm{cls}}(\delta)$. By $L$-smoothness of $L_{\mathrm{cls}}$,
\begin{align*}
L_{\mathrm{cls}}(\delta^{+})
&\leq
L_{\mathrm{cls}}(\delta)
+
a^\top(\delta^{+}-\delta)
+
\frac{L}{2}\|\delta^{+}-\delta\|_2^2 \\
&= L_{\mathrm{cls}}(\delta) - \eta a^\top G_\eta + \frac{L\eta^2}{2}\|G_\eta\|_2^2.
\end{align*}

Writing $a=d^\star-(d^\star-a)$ and using Eq.~\eqref{eq:projection_optimality}, we obtain
\begin{align*}
a^\top G_\eta
&=
(d^\star)^\top G_\eta
-
(d^\star-a)^\top G_\eta
\\
&\geq
\|G_\eta\|_2^2
-
\|d^\star-a\|_2\|G_\eta\|_2.
\end{align*}
Substituting this inequality into the smoothness bound yields
\begin{equation}
\label{eq:projected_descent_bound}
L_{\mathrm{cls}}(\delta^{+})
\leq
L_{\mathrm{cls}}(\delta)
-
\eta
\left[
\left(1-\frac{L\eta}{2}\right)\|G_\eta\|_2
-
\|d^\star-a\|_2
\right]
\|G_\eta\|_2.
\end{equation}

For our ellipsoidal direction $\|d^\star-a\|_2 = \gamma\xi\|a\|_2$, where $\xi\in[1/\sqrt{1+\lambda},1]$ as defined in Theorem~\ref{thm:ellip_decrease_main}. Therefore, the projected update strictly decreases $L_{\mathrm{cls}}$ whenever
\begin{equation}
\label{eq:projected_descent_condition}
\left(1-\frac{L\eta}{2}\right) \|G_\eta\|_2 > \gamma\xi\|a\|_2.
\end{equation}
Thus, projection changes the quantitative guarantee of Theorem~\ref{thm:ellip_decrease_main}, but a sufficient descent condition can still be established for the update used in practice. When the projection is inactive, $G_\eta=d^\star$ and the update reduces to the unprojected step analyzed in Theorem~\ref{thm:ellip_decrease_main}.

\paragraph{Empirical effect of projection.}
We further examine whether projection frequently prevents descent in practice. For this diagnostic experiment, we accept a projected update only if
\begin{equation}
L_{\mathrm{cls}}(\delta^{+}) < L_{\mathrm{cls}}(\delta),
\end{equation}
and otherwise discard the update. Table~\ref{tab:projected_descent} compares this descent-filtered variant with the original attack. Only $3.18\%$-$3.41\%$ of projected updates are rejected, while the ASR decreases by at most $0.38$ percentage points. This indicates that the projected update rarely fails to decrease $L_{\mathrm{cls}}$ in practice and has negligible impact on the overall attack effectiveness. Together with Eq.~\eqref{eq:projected_descent_condition}, these results clarify the relationship between the theoretical analysis and the projected update used in our implementation.

\begin{table*}[t]
\centering
\caption{Effect of enforcing descent after the $\ell_\infty$ projection. ASR is
reported in \%, and rejected updates denote the fraction of projected updates
that do not decrease $L_{\mathrm{cls}}$.}
\label{tab:projected_descent}

\begin{tabular}{lccccc}
\toprule
& \textbf{TENT} & \textbf{RPL} & \textbf{EATA} & \textbf{SAR} & \textbf{DeYO} \\
\midrule
Original ASR
& 89.59 & 93.17 & 90.21 & 95.90 & 91.15 \\
Filtered ASR
& 89.28 & 92.86 & 90.00 & 95.72 & 90.77 \\
Rejected updates (\%)
& 3.20 & 3.19 & 3.28 & 3.18 & 3.41 \\
\bottomrule
\end{tabular}

\end{table*}

\section{Additional Details on Experimental Setting} \label{appendix:imp_detail}

\paragraph{TTA method.} 
Following prior work \cite{wu2023uncovering, cong2024test, su2024adversarial, rifatadversarial}, we evaluate attacks under five BN-affine TTA methods: TENT \cite{wang2020tent} adapts batch-normalization parameters by minimizing entropy, RPL \cite{rusak2021if} uses robust pseudo-labels, EATA \cite{niu2022efficient} updates the model only on confident and stable samples, SAR \cite{niu2023towards} performs sharpness-aware minimization, DeYO \cite{lee2024entropy} regularizes adaptation through a distributional entropy.  

\paragraph{Implementation detail.}
We use the test batch size of 200. For attack generation, we set the number of outer iterations to $K=500$ and use $T=1$ batch per iteration. To improve robustness, at each iteration we randomly sample the data composition ratios from uniform distributions: the victim data ratio is sampled between 5\% and 20\%, the support data ratio between 40\% and 60\%, and the remainder consists of benign data. The step size for perturbation updates is set to $\eta=1.0$. For the priority-aware gradient alignment, we set the trust-radius parameter to $\gamma=0.5$ and use the misalignment scaling $\kappa$ of 10, 10, and 0.01 for CIFAR10-C, CIFAR100-C, and ImageNet-C, respectively. Our experiments are conducted on a server equipped with a 64-core AMD Ryzen Threadripper 3990X CPU @ 2.9GHz, 128 GB of RAM, and two NVIDIA RTX 4500 Ada GPUs. 

\paragraph{Adaptation of baseline.} Naively applying prior class-wise attacks \cite{su2024adversarial, rifatadversarial} to our sample-wise setting yields suboptimal performance, as their original optimization does not account for trigger-based selectivity. To ensure a fair and rigorous comparison, we adapt these methods by integrating their original loss functions into our proposed meta-learning framework (Section~\ref{sec:meta_attack}). This allows the baselines to be optimized over simulated adaptation tasks, enabling them to learn perturbations that generalize to trigger-carrying inputs effectively. We retain the exact loss formulations from their official implementations (RTTDP: \url{https://github.com/Gorilla-Lab-SCUT/RTTDP}, FCA: \url{https://github.com/Restuccia-Group/tta-adv}) to isolate the impact of our proposed alignment strategy versus their respective objectives.

\begin{figure}
\centering
   \begin{subfigure}{.2\linewidth}
   \includegraphics[width=\textwidth]{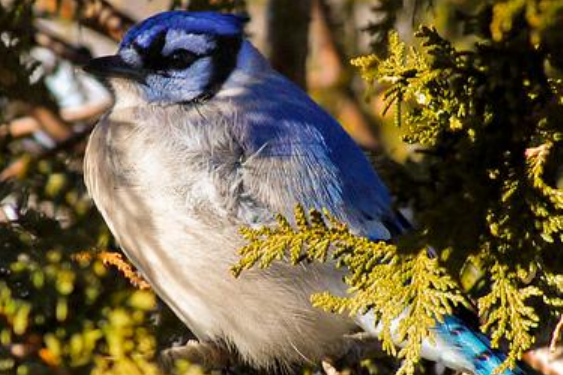}
   \caption{Reference.}
   \label{fig:testref}
   \end{subfigure}
   \begin{subfigure}{.2\linewidth}
   \includegraphics[width=\textwidth]{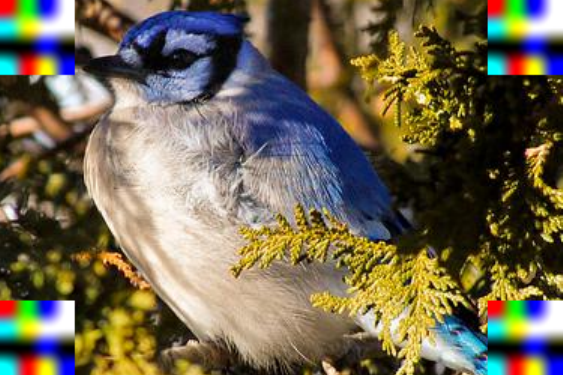}
   \caption{Patch.}
   \label{fig:testpatch}
   \end{subfigure}
   \begin{subfigure}{.2\linewidth}
   \includegraphics[width=\textwidth]{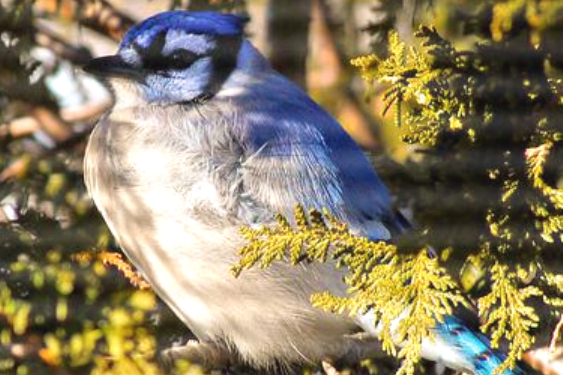}
   \caption{SIG.}
   \label{fig:testsig}
   \end{subfigure}
\caption{Trigger patterns on victim input.}
\label{fig:trigger_vis}
\end{figure}

\paragraph{Trigger specification.} We evaluate two common triggers in the experiment. For patch trigger \cite{gu2017badnets}, we use the trigger pattern at four corners, with the size of the trigger being 15\% of the image size. For SIG trigger \cite{barni2019new}, we superimpose a sinusoidal signal onto the entire image, parameterized by a frequency of 10 and an amplitude of $16/255$. The visualization is shown in Figure~\ref{fig:trigger_vis}. We also include an additional experiment using the imperceptible WaNet~\cite{nguyen2021wanet} trigger in Appendix~\ref{appendix:imperceptible_trigger}.

\paragraph{Evaluation protocol.} We follow \cite{su2024adversarial} and set the ratio of attacker-controlled samples to the combined victim and benign samples to $1:1$. Experiments with dynamically varying participation are provided in Appendix~\ref{appendix:random_participation}. We use two label-free metrics to quantify stealth on non-targeted inputs. EDD is the Wasserstein distance between their predictive-entropy distributions under benign and attacked adaptation. ADI measures the change in prediction disagreement between an input and its augmented version relative to benign adaptation.

\section{Additional Experiments} \label{appendix:more_exp}

\subsection{Attack Effectiveness under Random Attacker Participation} \label{appendix:random_participation}
\input{random}

We further evaluate a more dynamic setting in which the attacker-controlled ratio varies across batches and is sampled from $[0.1, 0.5]$ using either a uniform distribution or a Beta distribution skewed toward lower ratio (i.e., $0.1$), as shown in Tables~\ref{tab:random_ratio_uniform} and~\ref{tab:random_ratio_beta}, respectively. Under randomly varying attacker participation, our method remains substantially more effective than both baselines. With uniformly sampled attacker-controlled ratios, our attack achieves an average ASR of approximately $60\%$, compared with only about $9\%$ for RTTDP and $14\%$ for FCA. Even when the ratio is Beta-distributed and skewed toward the low-participation regime around $0.1$, our method maintains an average ASR of approximately $38\%$, whereas both baselines remain below $10\%$. This improvement does not come with a disproportionate increase in the label-free stealth metrics. Under uniform sampling, our entropy drift is comparable to FCA, while our augmentation inconsistency is consistently lower. Under the Beta-skewed setting, our method exhibits substantially lower entropy drift than FCA and comparable augmentation inconsistency, despite achieving roughly $4$-$5\times$ higher ASR. RTTDP yields lower values on both stealth metrics, but this is accompanied by substantially weaker attack effectiveness. Overall, these results demonstrate that our attack remains effective even when attacker participation varies across batches.

\subsection{Effectiveness with Imperceptible Triggers} \label{appendix:imperceptible_trigger}

The triggers used in our main evaluation (Section~\ref{sec:exp}) are representative choices rather than requirements of our attack. Our framework is compatible with a broad range of stealthy triggers developed in the data-poisoning literature~\cite{10102775}. To verify that the attack does not rely on visually salient triggers, we further evaluate it using WaNet~\cite{nguyen2021wanet}, a widely used imperceptible warping-based trigger.

\begin{table*}[t]
\centering
\caption{ASR (\%) of our attack using the imperceptible WaNet~\cite{nguyen2021wanet} trigger.}
\label{tab}
\begin{tabular}{ccccc}
\toprule
\textbf{TENT} & \textbf{RPL} & \textbf{EATA} & \textbf{SAR} & \textbf{DeYO} \\
\midrule

80.80 & 94.65 & 83.64 & 99.68 & 89.38 \\

\bottomrule
\end{tabular}
\end{table*}

As shown in Table~\ref{tab}, our attack remains highly effective across all TTA methods, achieving ASRs between $80.80\%$ and $99.68\%$. This demonstrates that the attack is not limited to the visible triggers considered in the main evaluation and remains effective with an imperceptible trigger.

More broadly, malicious perturbations may also resemble natural corruptions or distribution shifts, which are precisely the conditions under which TTA is commonly deployed. Consequently, manual inspection is neither reliable nor scalable, particularly for imperceptible triggers and high-volume online adaptation streams. This further motivates the need for automated defenses against such attacks.

\subsection{Additional Defenses} \label{appendix:more_def}

\subsubsection{MedBN TTA Defense \cite{park2024medbn}}

\begin{table*}[ht]
\tiny
\centering
\caption{ASR (\%) of our attack against the MedBN TTA defense \cite{park2024medbn} across different TTA methods on the CIFAR-10-C dataset, using SIG and patch triggers.}
\label{tab:medbn}
\resizebox{\linewidth}{!}{
\begin{tabular}{c|ccccc|ccccc|}
\toprule
\multirow{2}{*}{\textbf{Attack}} & \multicolumn{5}{c|}{\textbf{SIG Trigger}} & \multicolumn{5}{c|}{\textbf{Patch Trigger}} \\
\cmidrule(lr){2-11}
& TENT & RPL & EATA & SAR & DeYO & TENT & RPL & EATA & SAR & DeYO \\
\midrule
No attack & 2.85 & 3.07 & 3.10 & 3.04 & 2.85 & 2.77 & 3.16 & 3.02 & 3.64 & 2.97 \\
Ours & 89.59 & 93.17 & 90.21 & 95.90 & 91.15 & 79.28 & 86.78 & 80.34 & 93.33 & 82.48 \\
Ours + \cite{park2024medbn} & 78.03 & 76.92 & 79.15 & 84.44 & 79.14 & 71.85 & 81.29 & 73.82 & 83.06 & 73.93 \\
\bottomrule
\end{tabular}
}
\end{table*}

Table~\ref{tab:medbn} reports the full attack performance of our method against the MedBN TTA defense \cite{park2024medbn}.

\subsubsection{Trigger Purification Defense \cite{doan2020februus}}

\begin{table*}[ht]
\centering
\caption{ASR (\%) of our attack against the trigger purification defense \cite{doan2020februus} across different TTA methods on the CIFAR-10-C dataset, using patch triggers.}
\label{tab:februus}
\begin{tabular}{c|ccccc|}
\toprule
\multirow{2}{*}{\textbf{Attack}} & \multicolumn{5}{c|}{\textbf{Patch Trigger}} \\
\cmidrule(lr){2-6}
& TENT & RPL & EATA & SAR & DeYO \\
\midrule
No attack & 2.77 & 3.16 & 3.02 & 3.64 & 2.97 \\
Ours & 79.28 & 86.78 & 80.34 & 93.33 & 82.48 \\
Ours + \cite{doan2020februus} & 51.54 &	58.18 &	50.76 &	66.45 &	54.26 \\
\bottomrule
\end{tabular}
\end{table*}

We evaluate our attack against a trigger purification defense designed to mitigate backdoor attacks \cite{doan2020februus}. The results are presented in Table~\ref{tab:februus}. Although the defense reduces the ASR to some extent, our attack remains effective.

\subsection{Label Distribution Shift} \label{appendix:label_dist}

\begin{figure*}
    \centering
    \includegraphics[width=0.99\linewidth]{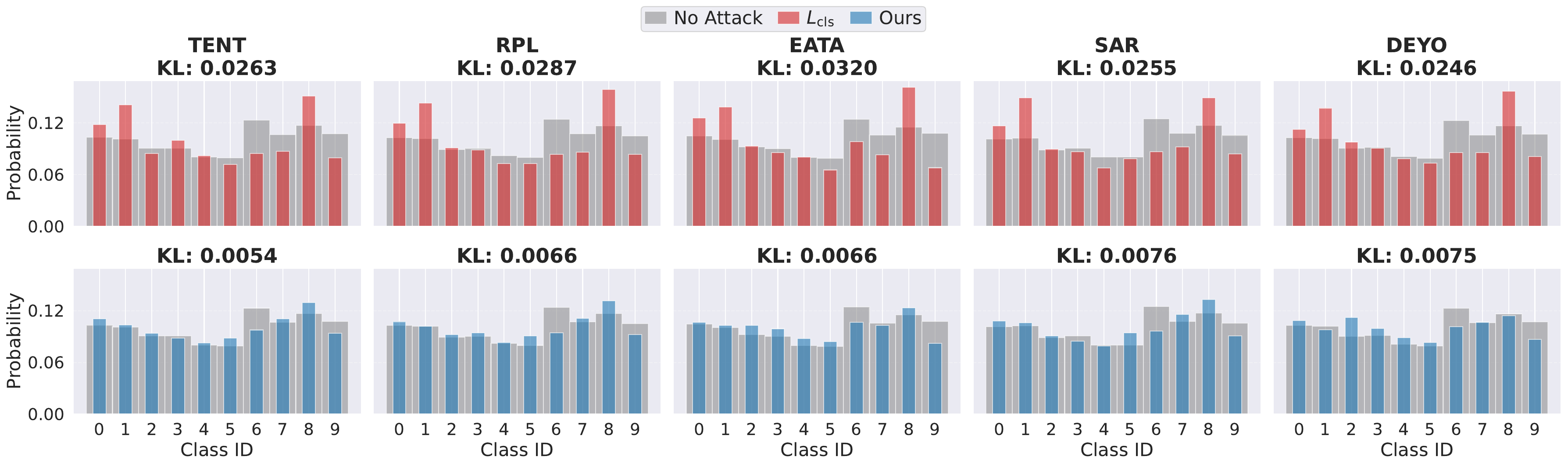}
    \caption{Comparison of label distribution shifts induced by $L_\mathrm{cls}$ and our method on CIFAR-10-C.}
    \label{fig:ce_10}
\end{figure*}

\begin{figure*}
    \centering
    \includegraphics[width=0.99\linewidth]{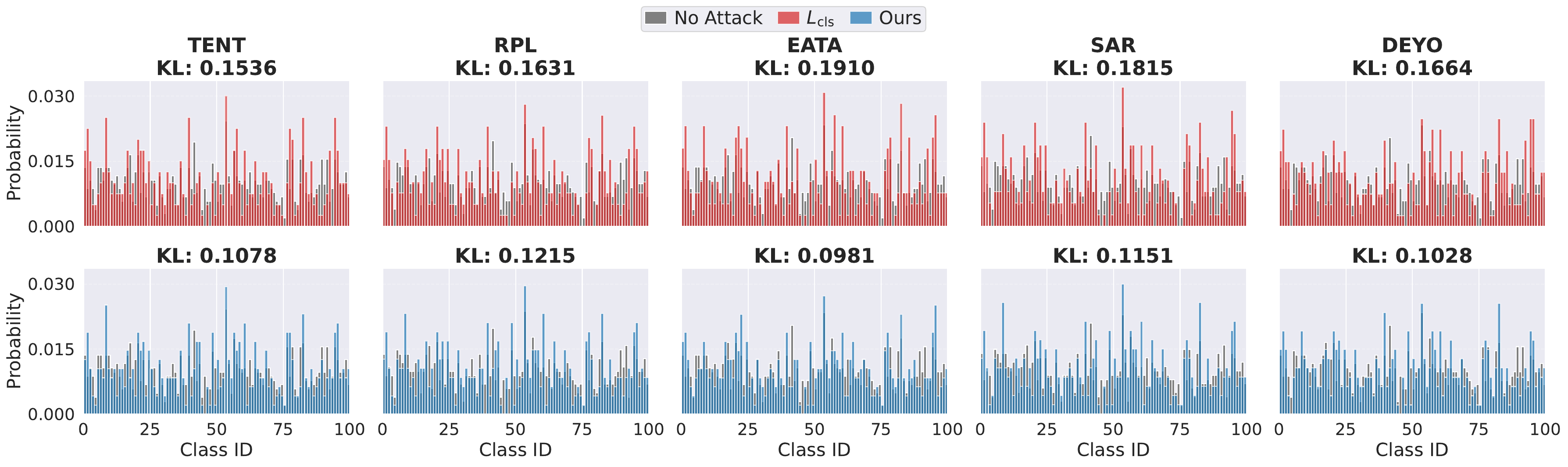}
    \caption{Comparison of label distribution shifts induced by $L_\mathrm{cls}$ and our method on CIFAR-100-C.}
    \label{fig:ce_100}
\end{figure*}

To complement the ablation study in Section~\ref{exp:abl}, Figures~\ref{fig:ce_10} and~\ref{fig:ce_100} visualize the label distribution shifts induced by attacks optimized using only the attack loss $L_{\mathrm{cls}}$. It exhibits substantially higher KL divergence with pronounced probability spikes compared to our approach.

\subsection{Sensitivity Analysis} \label{appendix:sen}

\begin{figure}[ht]
    \begin{subfigure}{.33\linewidth}
     \includegraphics[width = \textwidth]{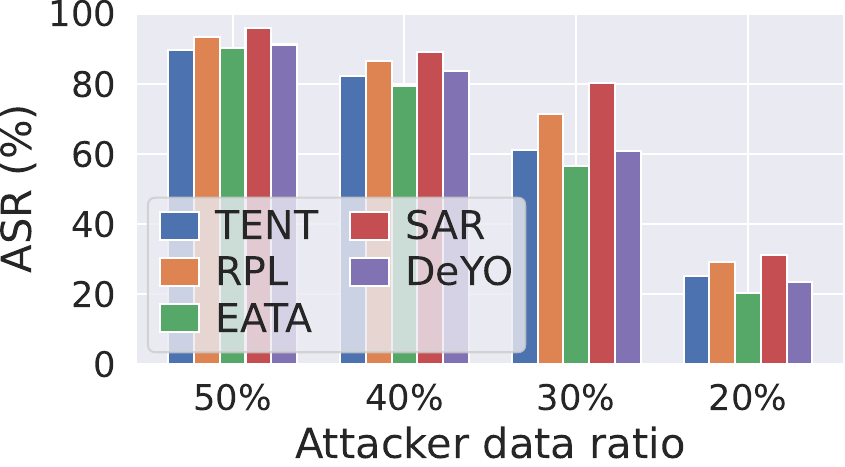}
     \caption{Attacker data ratio.}
     \label{fig:attacker_ratio}
    \end{subfigure}
    \begin{subfigure}{.33\linewidth}
     \includegraphics[width = \textwidth]{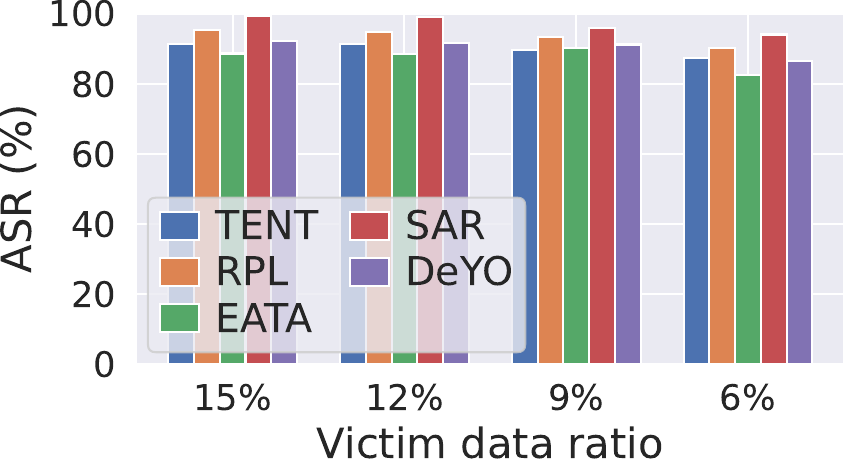}
     \caption{Victim data ratio.}
     \label{fig:victim_ratio}
    \end{subfigure}
    \begin{subfigure}{.33\linewidth}
     \includegraphics[width = \textwidth]{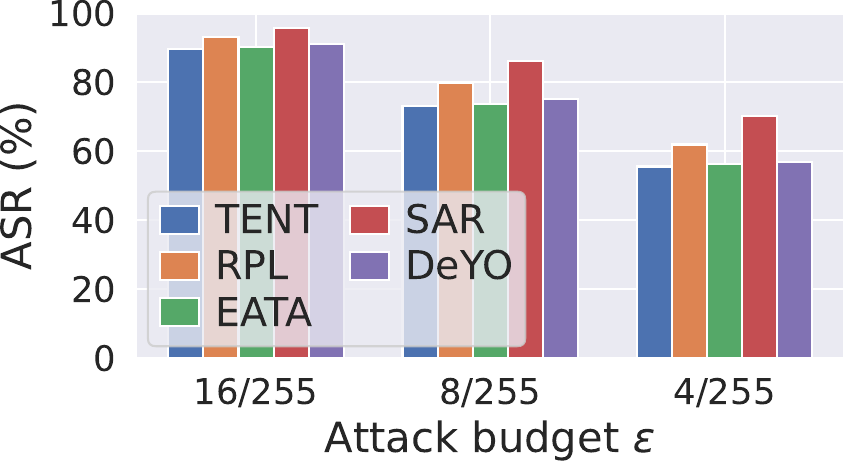}
     \caption{Attack budget.}
     \label{fig:attack_budget}
    \end{subfigure}
  \caption{Sensitivity analysis of the attack under different settings.}
  \label{fig:sensitivity}
\end{figure}

\begin{figure}
   \centering
   \includegraphics[width=0.50\linewidth]{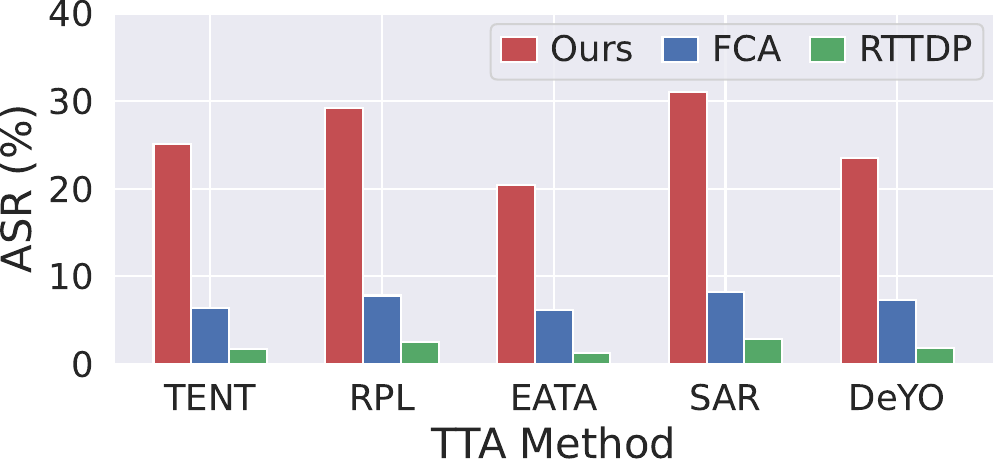}
   \caption{ASR (\%) across TTA methods at a 20\% attacker-controlled data ratio.}
   \label{fig:asr_at_20}
\end{figure}

\subsubsection{Dependency on Attacker Data Ratio} \label{appendix:ratio}

We evaluate the effectiveness of the attack on various attacker data ratios. The results are shown in Figure~\ref{fig:attacker_ratio}. Increasing the attacker data ratio consistently improves ASR across all TTA methods on CIFAR-10-C. At 50\%, the attack achieves the highest ASR, ranging from 89.59\% to 95.90\%. As the ratio decreases to 20\%, ASR declines. This behavior is expected, as increasing the proportion of attacker-controlled data within a batch makes it easier for the attacker to influence the batch statistics. A similar trend has been observed in prior TTA attacks \cite{wu2023uncovering, su2024adversarial, rifatadversarial}.

However, it is important to note that even at a 20\% attacker-controlled ratio, our attack remains substantially more effective than existing baselines, as shown in Figure~\ref{fig:asr_at_20}. On average, our attack achieves an ASR approximately $3.6\times$ higher than FCA and $12.5\times$ higher than RTTDP.

\subsubsection{Dependency on Victim Ratio}

We vary the victim ratio while keeping the batch size fixed, and report the ASR in Figure~\ref{fig:victim_ratio}. From the figure, increasing the victim ratio while keeping the batch size fixed generally leads to higher ASR. Intuitively, a larger victim ratio means more trigger-carrying victim samples in each batch and fewer unaffected benign inputs, since both the batch size and the number of attack-controlled samples remain fixed. As a result, the adaptation update is more consistently driven toward the malicious objective, making the attack easier to induce.

\subsubsection{Dependency on Attack Budget}

We further evaluate the attack's effectiveness on different attack budgets, with results reported in Figure~\ref{fig:attack_budget}. Specifically, we consider $\epsilon=16/255, \epsilon=8/255, \epsilon=4/255$. Increasing the attack budget $\epsilon$ consistently improves ASR across all TTA methods. With the largest budget of $16/255$, the attack achieves the highest ASR. As the budget decreases to $8/255$ and $4/255$, ASR correspondingly drops across all methods, indicating that stronger perturbation budgets lead to more effective attacks.

\subsection{Cross-Dataset Backbone} \label{appendix:cross}
\input{crossbackbone}

Table~\ref{tab:res32} and Table~\ref{tab:res50} show the attack effectiveness using ResNet-32 and ResNet-50 as a backbone across three datasets, respectively. The attack remains highly effective under both ResNet-32 and ResNet-50 backbones, achieving consistently high ASR across all settings. For the SIG trigger, ImageNet-C exhibits near-perfect ASR for both architectures, while CIFAR-10-C and CIFAR-100-C show comparatively lower but still substantial attack success. For the patch trigger, the attack generally achieves higher ASR on CIFAR-100-C and ImageNet-C than on CIFAR-10-C. Comparing architectures, ResNet-32 typically yields higher ASR than ResNet-50, suggesting that the attack transfers more effectively to the smaller backbone.

\subsection{Attack Effectiveness across Various Severities} \label{appendix:severities}

\input{severity}

Table~\ref{tab:sev_sig} summarizes attack effectiveness under varying corruption severities. For ImageNet-C, experiments are conducted only at severity level 3, as specified in Section~\ref{exp:setting}. Overall, ASR consistently increases as the corruption severity becomes stronger, indicating that heavier distribution shifts make TTA methods more vulnerable to the attack. On CIFAR-10-C and CIFAR-100-C, the attack achieves progressively higher ASR from Severity 1 to Severity 5 for all methods and trigger types. The SIG trigger generally performs better on CIFAR-10-C, while the patch trigger achieves particularly strong ASR on CIFAR-100-C. On ImageNet-C, the attack maintains very high ASR across all TTA methods for both trigger types.

\subsection{Attack Effectiveness across Various Corruption Types} \label{appendix:types}

\input{type_sig}

In Table~\ref{tab:type_sig}, we report the attack effectiveness across the 15 corruption types defined in~\cite{hendrycks2019robustness}. Stronger distribution-shift corruptions such as contrast, glass blur, Gaussian noise, shot noise, and JPEG compression generally lead to higher ASR, while brightness is consistently among the least vulnerable corruption types. On CIFAR-10-C, most corruption types yield high ASR, especially contrast, glass blur, fog, motion blur, and zoom blur. On CIFAR-100-C, the attack remains effective but shows larger variation across corruption types, with brightness and pixelate producing relatively lower ASR. For ImageNet-C, the ASR is near-saturated for many corruptions, including defocus blur, frost, Gaussian noise, glass blur, and shot noise, indicating that the attack remains highly effective under diverse corruption-induced distribution shifts.

\section{Limitation and Future Work} \label{appendix:limitation}

We acknowledge two limitations of our current study and outline corresponding future directions where applicable.
\paragraph{Scope of Stealth Evaluation.} We evaluate stealth using two label-free monitoring signals, EDD and ADI, which capture entropy-distribution drift and augmentation inconsistency relative to benign adaptation. While these metrics cover complementary aspects of prediction behavior, they do not encompass all possible runtime monitors, such as temporal drift detectors or feature-level anomaly detection. Evaluating against a broader range of label-free monitoring strategies is a direction for future work.
\paragraph{Performance under Very Low Attacker Participation.} While our attack outperforms existing baselines under dynamically varying attacker participation (Appendix~\ref{appendix:random_participation}) and under lower ratios (Appendix~\ref{appendix:ratio}), its effectiveness decreases as the attacker-controlled fraction becomes very small. Similar declines are also reported by prior TTA attacks~\cite{wu2023uncovering, su2024adversarial, rifatadversarial}, reflecting the reduced influence of fewer attacker-controlled samples on shared batch statistics and adaptation updates. Improving effectiveness under highly constrained injection budgets is a direction for future TTA attacks.

%% file: algo.tex
\begin{algorithm}
\caption{Pseudo-code of the attack generation}
\label{alg:attack}
\begin{algorithmic}[1]
\STATE \textbf{Input:} Model $f_{\theta_0}$; attacker data $X_A$ split into batches; target label $y_{\mathrm{tgt}}$; perturbation budget $\epsilon$; step size $\eta$; trust-radius parameter $\gamma$; misalignment scaling $\kappa$; number of outer iterations $K$; number of batches per iteration $T$.
\STATE \textbf{Output:} Perturbed data $\hat{X}_A := X_A + \delta_{N_A}$.

\STATE Initialize $\delta_{N_A} \leftarrow 0$\;
\STATE  \textbf{for} $k = 0,1,\dots,K-1$ \textbf{do}:
    \STATE \quad Sample batches $\{\tau_1,\dots,\tau_T\}$ from $X_A$
    \STATE \quad Initialize aggregated gradients $a \leftarrow 0$, $c \leftarrow 0$\;
    \STATE  \quad \textbf{for} each batch $\tau$ \textbf{do}:
        \STATE  \quad \quad Use $f_{\theta_0}$ to compute attack loss $L_{\mathrm{cls}}^{\tau}(\delta_{N_A})$ and stealth loss $L_{\mathrm{stl}}^{\tau}(\delta_{N_A})$

        \STATE  \quad \quad $a \leftarrow a + \nabla_\delta L_{\mathrm{cls}}^{\tau}(\delta_{N_A})$, \quad  $c \leftarrow c + \nabla_\delta L_{\mathrm{stl}}^{\tau}(\delta_{N_A})$
    
    \STATE \quad $a \leftarrow a / T$, \quad $c \leftarrow c / T$

    \STATE \quad $d^k \leftarrow \textsc{EllipsoidalDirection}(a, c, \gamma,\kappa)$ \# \textit{Eqs.~\eqref{eq:primal}--\eqref{eq:final_min}}

    \STATE \quad $\delta_{N_A} \leftarrow \Pi_{\mathcal{B}_\infty(\epsilon)}\bigl(\delta_{N_A} - \eta\, d^k\bigr)$ \# \textit{Projection and clipping}

\STATE \textbf{return} $\hat{X}_A := X_A + \delta_{N_A}$

\vspace{4pt}

\STATE \textbf{Subroutine} \textsc{EllipsoidalDirection}$(a,c,\gamma,\kappa)$:
\STATE \quad Construct $(u,\lambda)$ from $(a,c)$ (Appendix~\ref{appendix:metric_details})
\STATE \quad $\rho \leftarrow \gamma \|a\|_2$
\STATE \quad $g_w \leftarrow w a + (1-w)c$
\STATE \quad Solve $\min_{w\in[0,1]} a^\top g_w + \rho \|g_w\|_{M^{-1}}$
\STATE \quad \textbf{return} $d^\star(w^\star)$

\end{algorithmic}
\end{algorithm}

%% file: crossbackbone.tex
\begin{table*}[!htbp]
\centering
\caption{ASR (\%) of our attack using ResNet-32 backbone across TTA methods and datasets.}
\label{tab:res32}
\resizebox{\linewidth}{!}{
\begin{tabular}{l|rrrrr|rrrrr|}
\toprule
\multirow{2}{*}{\textbf{Dataset}} 
& \multicolumn{5}{c|}{\textbf{SIG}} 
& \multicolumn{5}{c|}{\textbf{Patch}} \\
\cmidrule(lr){2-6} \cmidrule(lr){7-11}
& TENT & RPL & EATA & SAR & DeYO 
& TENT & RPL & EATA & SAR & DeYO \\
\midrule
CIFAR-10-C  & 89.59 & 93.17 & 90.21 & 95.90 & 91.15 & 79.28 & 86.78 & 80.34 & 93.33 & 82.48 \\
CIFAR-100-C & 66.93 & 76.17 & 76.42 & 81.49 & 70.66 & 88.04 & 91.82 & 91.83 & 93.47 & 88.97 \\
ImageNet-C  & 99.07 & 99.28 & 99.05 & 99.23 & 98.15 & 91.59 & 91.76 & 91.51 & 91.57 & 90.74 \\
\bottomrule
\end{tabular}
}
\end{table*}

\begin{table*}[!htbp]
\centering
\caption{ASR (\%) of our attack using ResNet-50 backbone across TTA methods and datasets.}
\label{tab:res50}
\resizebox{\linewidth}{!}{
\begin{tabular}{l|rrrrr|rrrrr|}
\toprule
\multirow{2}{*}{\textbf{Dataset}} 
& \multicolumn{5}{c|}{\textbf{SIG}} 
& \multicolumn{5}{c|}{\textbf{Patch}} \\
\cmidrule(lr){2-6} \cmidrule(lr){7-11}
& TENT & RPL & EATA & SAR & DeYO 
& TENT & RPL & EATA & SAR & DeYO \\
\midrule
CIFAR-10-C  & 72.66 & 63.18 & 47.31 & 76.09 & 40.63 & 61.45 & 52.21 & 37.23 & 72.65 & 45.48 \\
CIFAR-100-C & 61.22 & 54.17 & 57.41 & 66.00 & 68.09 & 84.00 & 84.78 & 81.63 & 92.31 & 87.23 \\
ImageNet-C  & 96.79 & 96.99 & 95.22 & 96.98 & 95.49 & 92.29 & 92.49 & 88.48 & 92.28 & 89.21 \\
\bottomrule
\end{tabular}
}
\end{table*}

%% file: severity.tex
\begin{table*}[!htbp]
\tiny
\centering
\caption{ASR (\%) of our attack under various TTA methods and severities, with SIG and patch trigger.}
\label{tab:sev_sig}
\resizebox{\linewidth}{!}{
\begin{tabular}{cc|ccccc|ccccc|}
\toprule
\multirow{2}{*}{\textbf{Dataset}} & \multirow{2}{*}{{\textbf{Severity}}} & \multicolumn{5}{c|}{\textbf{SIG Trigger}} & \multicolumn{5}{c|}{\textbf{Patch Trigger}} \\
\cmidrule(lr){3-12}
& & {{TENT}} & {RPL} & {EATA} & {SAR} & {DeYO} & {{TENT}} & {RPL} & {EATA} & {SAR} & {DeYO} \\
\midrule
\multirow{5}{*}{\textbf{\begin{tabular}[c]{@{}c@{}}CIFAR-10-C \\ (ResNet32)\end{tabular}}} 
& Severity 1 & 87.33 & 90.46 & 88.27 & 94.29 & 88.70 & 72.12 & 80.62 & 73.60 & 89.73 & 75.39 \\
& Severity 2 & 89.10 & 92.24 & 89.27 & 95.30 & 90.17 & 76.73 & 84.45 & 76.95 & 91.64 & 79.79 \\
& Severity 3 & 89.72 & 93.78 & 89.66 & 95.89 & 91.18 & 79.40 & 87.54 & 80.35 & 93.23 & 83.03 \\
& Severity 4 & 90.54 & 94.47 & 91.53 & 96.85 & 92.26 & 82.21 & 89.27 & 83.28 & 95.41 & 85.24 \\
& Severity 5 & 91.28 & 94.87 & 92.33 & 97.15 & 93.46 & 85.95 & 92.05 & 87.53 & 96.67 & 88.95 \\
\midrule
\multirow{5}{*}{\textbf{\begin{tabular}[c]{@{}c@{}}CIFAR-100-C \\ (ResNet32)\end{tabular}}}  
& Severity 1 & 55.35 & 64.62 & 63.58 & 70.96 & 58.45 & 82.32 & 87.74 & 87.83 & 89.97 & 83.51 \\
& Severity 2 & 61.98 & 72.37 & 72.81 & 79.39 & 66.47 & 85.86 & 90.88 & 90.28 & 92.93 & 87.11 \\
& Severity 3 & 67.69 & 76.48 & 77.57 & 81.95 & 70.59 & 88.62 & 91.76 & 92.03 & 93.64 & 89.30 \\
& Severity 4 & 72.98 & 81.75 & 82.73 & 86.54 & 77.54 & 90.84 & 93.90 & 94.34 & 95.23 & 91.83 \\
& Severity 5 & 76.65 & 85.65 & 85.41 & 88.62 & 80.26 & 92.59 & 94.79 & 94.69 & 95.55 & 93.10 \\
\midrule
\textbf{\begin{tabular}[c]{@{}c@{}}ImageNet-C \\ (ResNet50)\end{tabular}}
& Severity 3 & 96.79 & 96.99 & 95.22 & 96.98 & 95.49 & 92.29 & 92.49 & 88.48 & 92.28 & 89.21 \\
\bottomrule

\end{tabular}
}
\end{table*}

%% file: type_sig.tex
\begin{table*}[!htbp]
\centering
\caption{ASR (\%) of our attack under various TTA methods and corruption types, with SIG and patch trigger.}
\label{tab:type_sig}
\resizebox{\linewidth}{!}{
\begin{tabular}{cc|ccccc|ccccc|}
\toprule
\multirow{2}{*}{\textbf{Dataset}} & \multirow{2}{*}{{\textbf{Corruption Type}}} & \multicolumn{5}{c|}{\textbf{SIG Trigger}} & \multicolumn{5}{c|}{\textbf{Patch Trigger}} \\
\cmidrule(lr){3-12}
& & {{TENT}} & {RPL} & {EATA} & {SAR} & {DeYO} & {{TENT}} & {RPL} & {EATA} & {SAR} & {DeYO} \\
\midrule
\multirow{15}{*}{\textbf{\begin{tabular}[c]{@{}c@{}}CIFAR-10-C \\ (ResNet32)\end{tabular}}} 
& Brightness & 75.84 & 81.33 & 77.22 & 85.26 & 77.66 & 60.55 & 69.07 & 62.34 & 79.74 & 63.93 \\
& Contrast & 97.15 & 98.06 & 97.31 & 98.70 & 96.80 & 85.22 & 89.58 & 84.74 & 90.84 & 84.47 \\
& Defocus Blur & 91.45 & 92.71 & 91.87 & 95.27 & 91.88 & 80.61 & 85.18 & 80.71 & 90.71 & 82.98 \\
& Elastic Transform & 92.55 & 95.26 & 92.50 & 96.89 & 93.46 & 84.38 & 90.46 & 84.03 & 94.57 & 86.52 \\
& Fog & 95.47 & 97.35 & 95.44 & 97.86 & 94.98 & 80.76 & 86.99 & 81.70 & 90.58 & 82.70 \\
& Frost & 86.55 & 90.79 & 89.56 & 94.41 & 89.63 & 69.50 & 78.34 & 72.53 & 91.54 & 73.58 \\
& Gaussian Noise & 89.29 & 94.26 & 88.87 & 97.33 & 91.12 & 77.25 & 89.59 & 78.07 & 96.58 & 81.87 \\
& Glass Blur & 94.42 & 97.72 & 94.92 & 98.99 & 96.33 & 87.83 & 94.66 & 88.90 & 97.35 & 91.57 \\
& Impulse Noise & 86.80 & 93.81 & 87.94 & 96.85 & 90.50 & 75.07 & 87.30 & 77.36 & 96.50 & 82.05 \\
& Jpeg Compression & 88.88 & 92.96 & 90.14 & 96.70 & 91.19 & 89.51 & 95.16 & 90.26 & 98.76 & 91.91 \\
& Motion Blur & 94.27 & 96.50 & 93.50 & 97.52 & 95.08 & 85.99 & 90.66 & 86.06 & 94.35 & 87.33 \\
& Pixelate & 86.76 & 91.99 & 87.58 & 95.18 & 89.11 & 83.85 & 89.43 & 83.85 & 95.41 & 86.13 \\
& Shot Noise & 86.41 & 91.16 & 86.81 & 96.40 & 87.83 & 73.31 & 85.26 & 74.51 & 95.43 & 77.97 \\
& Snow & 86.37 & 89.19 & 87.59 & 94.09 & 89.12 & 69.17 & 79.24 & 73.01 & 93.04 & 75.57 \\
& Zoom Blur & 91.70 & 94.38 & 91.94 & 97.01 & 92.62 & 86.23 & 90.86 & 87.05 & 94.61 & 88.62 \\
\midrule
\multirow{15}{*}{\textbf{\begin{tabular}[c]{@{}c@{}}CIFAR-100-C \\ (ResNet32)\end{tabular}}}  
& Brightness & 45.36 & 51.22 & 51.64 & 57.34 & 47.73 & 67.84 & 73.98 & 74.41 & 78.40 & 68.49 \\
& Contrast & 76.44 & 86.51 & 84.04 & 89.89 & 81.17 & 91.95 & 93.43 & 92.92 & 93.99 & 91.80 \\
& Defocus Blur & 61.06 & 67.84 & 66.18 & 73.84 & 63.81 & 84.30 & 88.04 & 87.47 & 90.27 & 86.01 \\
& Elastic Transform & 66.89 & 75.12 & 75.16 & 81.03 & 69.40 & 88.20 & 92.69 & 91.32 & 93.55 & 88.77 \\
& Fog & 74.41 & 81.98 & 84.72 & 88.26 & 78.94 & 91.19 & 93.14 & 94.27 & 95.49 & 91.91 \\
& Frost & 63.25 & 73.17 & 75.92 & 80.12 & 69.21 & 81.66 & 87.47 & 87.65 & 88.87 & 84.09 \\
& Gaussian Noise & 73.93 & 85.51 & 87.07 & 89.87 & 80.00 & 92.33 & 97.25 & 97.48 & 97.70 & 93.22 \\
& Glass Blur & 72.42 & 81.85 & 83.12 & 86.72 & 76.22 & 95.20 & 97.90 & 98.28 & 98.69 & 96.08 \\
& Impulse Noise & 74.95 & 84.69 & 84.16 & 88.21 & 78.50 & 89.78 & 94.73 & 94.26 & 96.22 & 91.75 \\
& Jpeg Compression & 71.52 & 82.03 & 83.37 & 86.73 & 74.53 & 93.50 & 96.46 & 97.51 & 98.08 & 94.24 \\
& Motion Blur & 69.09 & 76.30 & 76.59 & 80.75 & 71.40 & 90.71 & 93.02 & 92.34 & 94.18 & 90.90 \\
& Pixelate & 56.28 & 65.05 & 64.92 & 73.47 & 59.02 & 86.60 & 90.29 & 90.87 & 92.61 & 86.19 \\
& Shot Noise & 69.10 & 83.00 & 81.95 & 86.03 & 73.51 & 91.51 & 96.32 & 96.26 & 97.62 & 93.11 \\
& Snow & 61.49 & 69.27 & 69.80 & 76.59 & 64.61 & 84.71 & 89.05 & 89.26 & 91.60 & 86.27 \\
& Zoom Blur & 67.78 & 79.07 & 77.64 & 83.55 & 71.89 & 91.21 & 93.48 & 93.22 & 94.71 & 91.70 \\
\midrule
\multirow{15}{*}{\textbf{\begin{tabular}[c]{@{}c@{}}ImageNet-C \\ (ResNet50)\end{tabular}}}  
& Brightness & 84.02 & 84.62 & 82.84 & 84.02 & 84.52 & 72.19 & 72.19 & 66.86 & 71.60 & 66.07 \\
& Contrast & 100.00 & 100.00 & 99.35 & 100.00 & 99.36 & 94.59 & 94.59 & 92.26 & 94.59 & 92.95 \\
& Defocus Blur & 100.00 & 100.00 & 100.00 & 100.00 & 100.00 & 100.00 & 100.00 & 97.94 & 100.00 & 97.89 \\
& Elastic Transform & 92.36 & 92.36 & 88.61 & 92.99 & 88.20 & 85.99 & 86.62 & 77.85 & 85.99 & 78.40 \\
& Fog & 97.89 & 98.59 & 97.95 & 98.59 & 97.95 & 85.21 & 85.21 & 84.93 & 85.21 & 84.35 \\
& Frost & 100.00 & 100.00 & 98.94 & 100.00 & 98.92 & 100.00 & 100.00 & 96.81 & 100.00 & 96.81 \\
& Gaussian Noise & 100.00 & 100.00 & 98.10 & 100.00 & 98.15 & 99.01 & 98.98 & 97.14 & 98.99 & 97.22 \\
& Glass Blur & 100.00 & 100.00 & 100.00 & 100.00 & 100.00 & 100.00 & 100.00 & 100.00 & 100.00 & 100.00 \\
& Impulse Noise & 98.15 & 98.15 & 98.25 & 98.15 & 98.25 & 98.15 & 98.13 & 93.86 & 98.15 & 96.49 \\
& Jpeg Compression & 93.38 & 93.38 & 86.99 & 94.12 & 89.73 & 83.09 & 83.09 & 77.40 & 83.09 & 79.45 \\
& Motion Blur & 95.04 & 95.87 & 93.39 & 95.87 & 93.50 & 95.04 & 95.04 & 88.43 & 95.04 & 91.06 \\
& Pixelate & 92.76 & 92.76 & 90.85 & 92.76 & 91.50 & 80.92 & 81.58 & 74.34 & 79.61 & 75.32 \\
& Shot Noise & 100.00 & 100.00 & 99.12 & 100.00 & 100.00 & 100.00 & 100.00 & 97.37 & 100.00 & 98.25 \\
& Snow & 99.06 & 99.05 & 94.78 & 99.06 & 93.91 & 95.28 & 96.19 & 87.83 & 96.23 & 89.47 \\
& Zoom Blur & 99.15 & 100.00 & 99.16 & 99.15 & 98.37 & 94.92 & 95.73 & 94.17 & 95.73 & 94.35 \\
\bottomrule

\end{tabular}
}
\end{table*}